\documentclass[10pt,twocolumn,letterpaper]{article}

\usepackage[pagenumbers]{cvpr} 

\definecolor{cvprblue}{rgb}{0.21,0.49,0.74}
\usepackage[pagebackref,breaklinks,colorlinks,allcolors=cvprblue]{hyperref}

\usepackage{url}

\usepackage[utf8]{inputenc} 
\usepackage[T1]{fontenc}    
\usepackage{hyperref}       
\usepackage{url}            
\usepackage{booktabs}       
\usepackage{amsfonts}       
\usepackage{nicefrac}       
\usepackage{microtype}      
\usepackage{xcolor}         
\usepackage{graphicx}
\usepackage{subcaption}
\usepackage{wrapfig}
\usepackage{amsmath}
\usepackage{amssymb}
\usepackage[export]{adjustbox}
\usepackage{arydshln}

\usepackage{easymath}
\usepackage{multirow}
\usepackage{makecell}
\usepackage{verbatim}
\usepackage{colortbl}
\usepackage{color} 
\usepackage{tcolorbox} 
\usepackage{algorithm}
\usepackage{algpseudocode}
\usepackage{threeparttable}
\usepackage[absolute,overlay]{textpos} 
\usepackage{xcolor}
\definecolor{forestgreen}{RGB}{34,139,34}

\usepackage[capitalize]{cleveref}
\crefname{theorem}{Theorem}{Theorems}
\Crefname{theorem}{Theorem}{Theorems}
\crefname{algorithm}{Alg.}{Algs.}
\Crefname{algorithm}{Alg.}{Algs.}
\crefname{figure}{Fig.}{Figs.}
\Crefname{figure}{Fig.}{Figs.}
\crefname{equation}{Eq.}{Eq.}

\usepackage{siunitx}
\definecolor{oracleCol}{RGB}{235,235,235}
\definecolor{ilstopCol}{RGB}{225,240,255}
\newcommand{\imgbox}[1]{%
    \setlength{\fboxsep}{0pt}
    \setlength{\fboxrule}{0.8pt}
    \fcolorbox{white}{white}{\includegraphics[width=0.15\textwidth]{#1}}%
}

\newcommand{\imgboxsel}[1]{%
    \setlength{\fboxsep}{0pt}%
    \setlength{\fboxrule}{0.8pt}%
    \fcolorbox{blue}{white}{\includegraphics[width=0.15\textwidth]{#1}}%
}
\usepackage{amsthm}

\def\paperID{5320} 
\def\confName{CVPR}
\def\confYear{2026}

\title{PR-MaGIC: Prompt Refinement Via Mask Decoder Gradient Flow For In-Context Segmentation}

\author{
Minjae Lee\thanks{Equal contribution.},
Sungwoo Hur\footnotemark[1],
Soojin Hwang,
Won Hwa Kim\\
Pohang University of Science and Technology, Pohang, South Korea\\
{\small \{lalswo010,hursungwoo,soojin0622,wonhwa\}@postech.ac.kr} \\
{\small \url{https://postech-minjaelee.github.io/PR-MaGIC/}}
\vspace{-5pt}
}

\newtheorem{proposition}{Proposition} 

\def \ourmodel{PR-MaGIC}

\begin{document}
\maketitle
\begin{abstract}
Visual Foundation Models (VFMs) such as the Segment Anything Model (SAM) have significantly advanced broad use of image segmentation. 
However, SAM and its variants necessitate substantial manual effort for prompt generation and additional training for specific applications. 
Recent approaches address these limitations by integrating SAM into in-context (one/few shot) segmentation, enabling auto-prompting through semantic alignment between query and support images. 
Despite these efforts, they still generate sub-optimal prompts that degrade segmentation quality due to visual inconsistencies between support and query images. 
To tackle this limitation, we introduce \ourmodel{} (\textbf{P}rompt \textbf{R}efinement via \textbf{Ma}sk Decoder \textbf{G}radient Flow for \textbf{I}n-\textbf{C}ontext Segmentation), a training-free test-time framework that refines prompts via gradient flow derived from SAM’s mask decoder. \ourmodel{} seamlessly integrates into in-context segmentation frameworks, being theoretically grounded yet practically stabilized through a simple top-1 selection strategy that ensures robust performance across samples.
Extensive evaluations demonstrate that \ourmodel{} consistently improves segmentation quality across various benchmarks, effectively mitigating inadequate prompts without requiring additional training or architectural modifications.
\end{abstract}    
\vspace{-12pt}
\section{Introduction}
\label{sec:intro}
\vspace{-5pt}

Large models have paved a new era of foundation models, e.g., Large Language Models (LLMs) \citep{zeng2022glm,touvron2023llama,brown2020language}, demonstrating exceptional versatility and generality across various tasks.  
In computer vision, Visual Foundation Models (VFMs) are transforming tasks such as image classification \citep{dosovitskiy2021image, touvron2021training}, object detection \citep{carion2020end, liu2021swin}, and segmentation \citep{cheng2021per, strudel2021segmenter} by leveraging large-scale data and advanced architectures.
Specifically in segmentation, the Segment Anything Model (SAM) \citep{kirillov2023segment} addresses the challenges of segmenting objects with diverse appearance by taking prompts such as points, bounding boxes, and coarse masks.

\begin{figure}[!t]
    \centering
    \centering
    
    \includegraphics[width=0.8\linewidth]{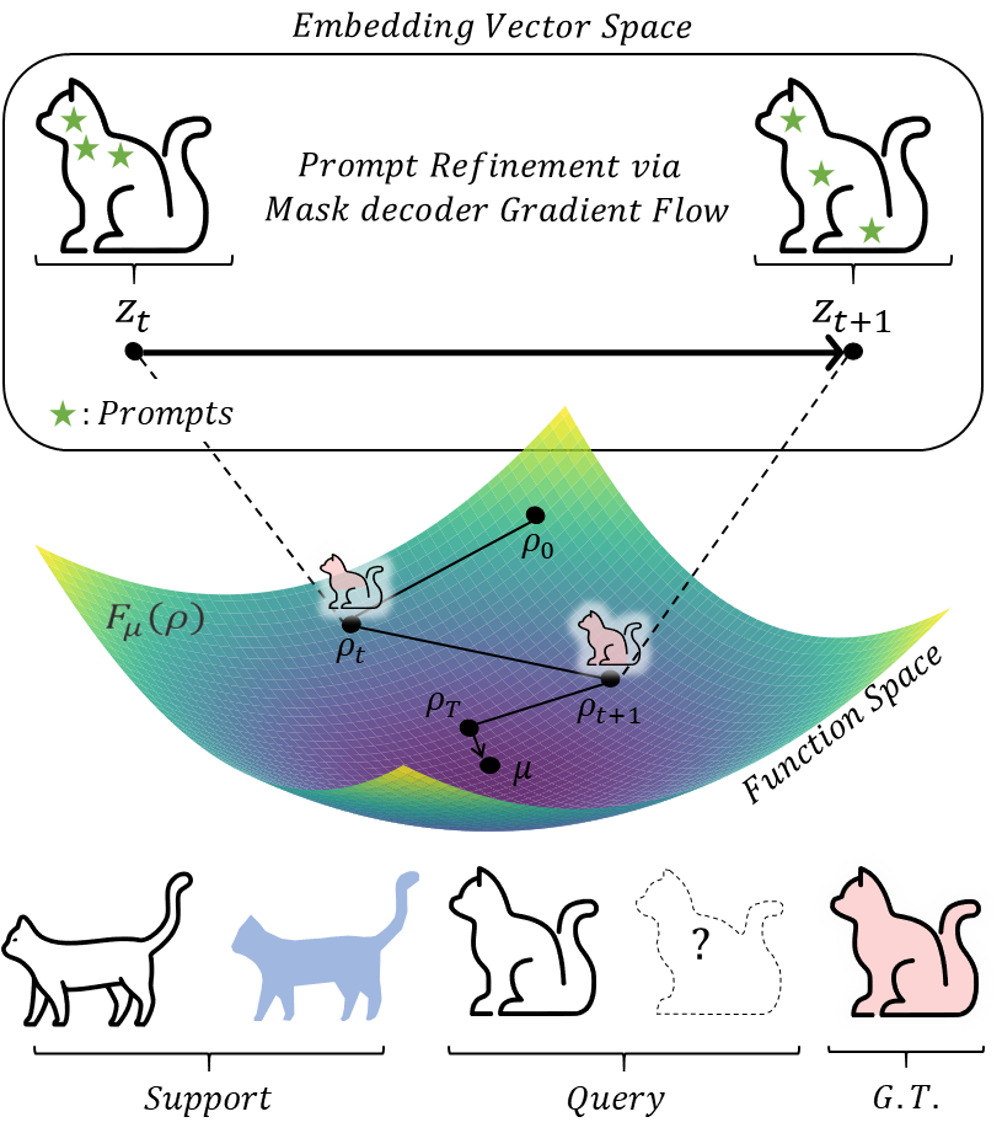}
    \vspace{-5pt}
    \caption{\textbf{Illustration of \ourmodel{}.} 
    \ourmodel{} iteratively refines prompts for segmentation by updating the embedding vector distribution $\rho_t$ with mask decoder gradient flow. 
    This process minimizes the KL divergence between $\rho_t$ and the target embedding vector distribution $\mu$ 
    from the ground truth mask. 
    At each iteration $t$, given an initial set of prompts and their corresponding segmentation masks, the embedding vector $z_t$ is updated along the gradient flow derived from the mask decoder, which in turn updates $\rho_t$. 
    }
    \label{fig:overview_intro}
    \vspace{-20pt}
\end{figure}

\begin{figure*}[!ht]
    \centering
    \begin{subfigure}[b]{0.19\textwidth}
        \includegraphics[width=\linewidth]{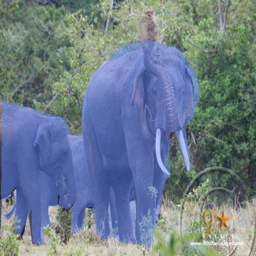}
         \caption{Support Image}
    \end{subfigure}
    \begin{subfigure}[b]{0.19\textwidth}
        \includegraphics[width=\linewidth]{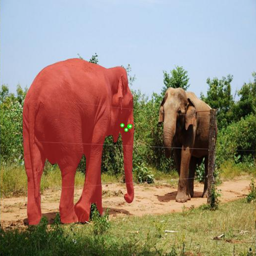}
        \caption{PerSAM-F}
        \label{fig:intro_b}
    \end{subfigure}
    \begin{subfigure}[b]{0.19\textwidth}
        \includegraphics[width=\linewidth]{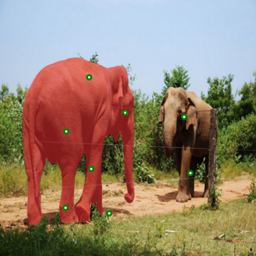}
        \caption{Matcher}
        \label{fig:intro_c}
    \end{subfigure}
    \begin{subfigure}[b]{0.19\textwidth}
        \includegraphics[width=\linewidth]{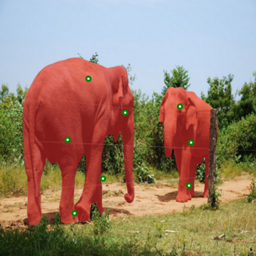}
        \caption{\ourmodel}
        \label{fig:intro_d}
    \end{subfigure}
    \begin{subfigure}[b]{0.19\textwidth}
        \includegraphics[width=\linewidth]{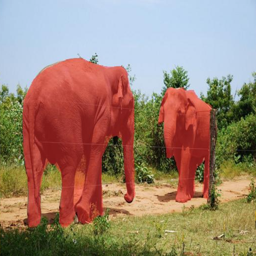}
        \caption{Ground Truth}
        \label{fig:intro_e}
    \end{subfigure}

    \vspace{-5pt}
    \caption{ \textbf{Example of prompts (green dots) and segmentation results (red masks).}
    (a) Support Image (elephant), 
    (b) and (c) Misaligned prompts and segmentation results from PerSAM-F and Matcher, 
    (d) Result from Matcher refined by \ourmodel{}, 
    (e) Ground Truth.
    }
    \label{fig:intro}
    \vspace{-10pt}
\end{figure*}

While SAM is highly versatile, it often produces imprecise boundaries and small holes under complex conditions, as it was trained for general object segmentation at relatively low resolution. Initial efforts fine-tuned SAM to improve segmentation precision across various scenarios \citep{chen2023sam, ke2024segment, wu2023medical}. However, these methods still substantially depend on additional parameters, extra curated data, and manual effort to obtain suitable prompts.
Recent works integrate VFMs into one/few-shot (in-context) segmentation to enable auto-prompting via support–query semantic alignment. PerSAM~\citep{zhang2024personalize} uses a single support image with a rough mask to build similarity-based prompts, and PerSAM-F~\citep{zhang2024personalize} further fine-tunes SAM to linearly combine hierarchical masks to reduce ambiguity. Matcher~\citep{liu2024matcher} assembles existing VFMs with sophisticated prompt sampling to solve one/few-shot segmentation without extra training.


Although the methods mentioned above have been successful, variations in appearance (e.g., color, viewpoint, and shape) between query and support images often lead to misleading similarity maps. This causes auto-prompting segmentation frameworks (e.g., PerSAM-F~\citep{zhang2024personalize} and Matcher~\citep{liu2024matcher}) to generate inadequate prompts, e.g., false-positive, semantically ambiguous, or semantically insufficient prompts, that confuse the SAM's decoder in capturing the target object, thereby compromising the quality of the segmentation mask.

This challenge motivates us to explore a refinement strategy that updates the initial suboptimal prompts produced by existing auto-prompting segmentation frameworks using mask decoder gradient flow. 
Such an approach leverages the rich information embedded in the decoder, which has been jointly trained with the encoder during SAM’s large-scale pretraining. 
In this regime, we propose \textbf{P}rompt \textbf{R}efinement via \textbf{Ma}sk Decoder \textbf{G}radient Flow for \textbf{I}n-\textbf{C}ontext Segmentation (\ourmodel{}). 
\ourmodel{} is a \emph{training-free}, inference-time framework that updates the query embedding via a mask–decoder–driven gradient flow, resamples prompts, and thereby improves the segmentation quality. 
The update is motivated by 
finding the distribution that yields the highest mask quality under the VFM’s mask decoder, which aligns the query embedding with the target object’s semantics during inference. \cref{fig:overview_intro} overviews the gradient flow, where the query embedding distribution is progressively refined. 

Our theoretical analysis shows exponential convergence of the gradient flow \emph{under a proximity assumption}: the initial embedding distribution lies near a decoder-optimal neighborhood. 
For the initial auto-prompts that do not comply with the assumption, 
we explore multiple refinement steps and adaptively select the most reliable mask to achieve stable performance across samples.
The key contributions of this work are as follows:
\begin{itemize}
\item \textbf{Training-free test-time refinement via gradient flow.} We introduce a novel, training-free refinement method that updates the query embedding with a mask–decoder–driven gradient flow and resamples prompts, improving segmentation quality without learnable parameters, architectural changes, or extra data.

\item \textbf{Segmentation mask selection for robustness.} 
We generate candidate masks across $T$ iterations and propose a simple top-1 support–query similarity selection to choose the final mask, which safeguards against step-size sensitivity and sample-dependent instability.


\item \textbf{Convergence analysis under a proximity assumption.} We provide a theoretical analysis showing that, \emph{when the initial embedding distribution is near a decoder-optimal neighborhood}, the entropy-regularized KL gradient flow drives exponential convergence of the query embeddings. We further complement this with a sensitivity study indicating that the assumption can be fragile in practice, motivating our segmentation mask selection step.
\end{itemize}

We validate \ourmodel{} on six datasets and two tasks (semantic and part segmentation), demonstrating consistent gains as a plug-and-play module, particularly when the semantic gap between support and query is not excessive.
A comparative exemplar result is given in \cref{fig:intro}, which illustrates that \ourmodel{} produces better prompts and more precise masks successfully capturing the complete target objects, unlike PerSAM-F and Matcher which tend to yield less effective prompts. 

\vspace{-5pt}
\section{Related Works}
\vspace{-5pt}

\subsection{One/Few-shot segmentation}

Few-shot segmentation aims to develop models capable of segmenting new classes using only a few annotated samples \citep{shaban2017one}, making it particularly useful in scenarios where acquiring large annotated data is difficult. Typically, these models use a pre-trained feature extractor to process both support and query images, combining information from support images 
to perform segmentation \citep{zhang2022feature,hong2022cost}. 
    
Recently, several approaches leveraging SAM as a pre-trained model have been proposed. PerSAM relies solely on one-shot data, comprising a support image and a rough mask of the target object. It utilizes a similarity map between a query and target object embeddings to locate the target object precisely, facilitating prompt generation. However, PerSAM faces challenges when there is large visual ambiguity, such as objects with visually distinct subparts or hierarchical structures \citep{zhang2024personalize}.          
PerSAM-F, fine-tuning variant of PerSAM, addresses this problem by introducing learnable parameters to fine-tune SAM, producing a linear combination of potential segmentation masks of different hierarchical levels to resolve ambiguity \citep{zhang2024personalize}. This approach improves segmentation accuracy and prevents overfitting on one-shot data, demonstrating enhanced performance in challenging scenarios.    
Similarly, Matcher \citep{liu2024matcher}, effectively combines pre-existing VFMs to tackle one/few-shot segmentation tasks without additional training. Matcher utilizes DINOv2 \citep{oquab2023dinov2} as an encoder to compute a similarity map with semantic understanding to accurately identify the location of a target object in the query image. It then employs SAM as a segmenter to obtain segmentation masks, benefiting from the zero-shot segmentation performance of the SAM. 

\subsection{Fine-tuning variants of SAM}

For efficient adaptation of SAM to various downstream tasks, a range of fine-tuning variants has emerged. HQ-SAM~\citep{ke2024segment}, for instance, adds a high-quality output token and trains on the HQSeg-44K dataset to improve mask precision. VRP-SAM~\citep{sun2024vrp} implements a visual reference prompt encoder, requiring specific reference images and detailed training. SAM-Adapter~\citep{chen2023sam} modifies the architecture by adding lightweight adapter layers while keeping most of SAM's original parameters frozen, reducing training effort. MobileSAM~\citep{zhang2023faster} replaces the heavy ViT-H encoder with Tiny-ViT, reducing the model size and complexity while still requiring adaptation.

All of these methods, however, necessitate additional training data and structural changes to the model architecture. These fine-tuning strategies still heavily depend on manual prompts, with training datasets requiring carefully crafted prompt annotations alongside segmentation data. This reliance on curated prompts limits their applicability in real-world settings, where obtaining accurate prompts and consistent user interaction may be challenging.

\section{Method}

\begin{figure*}[!t]
  \centering
  \includegraphics[width=\textwidth]{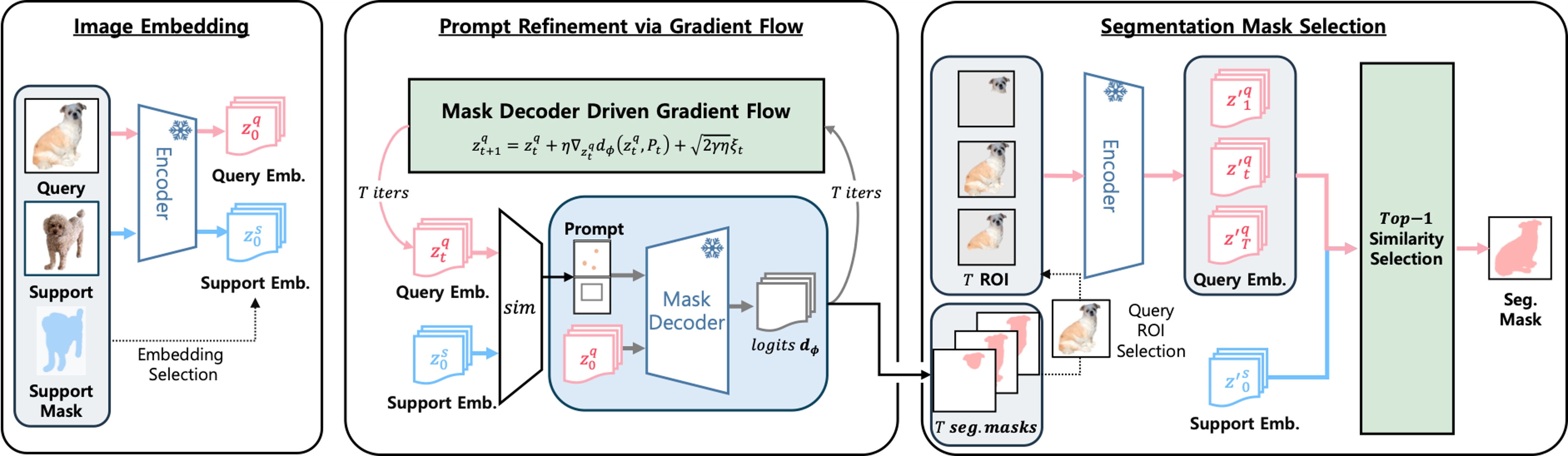}


  \caption{{\bf Overview of \ourmodel{}}. \emph{Encoder} box denotes image encoder ($E_\theta$), \emph{sim} box denotes similarity computing module, and \emph{Mask Decoder} box denotes SAM's mask decoder. 
  SAM's prompt encoder is omitted here to show the clear flow of our method, and green boxes denote our core method.
    \textbf{1) Image Embedding}: \emph{Encoder} maps the support and query images to embeddings $\mathbf{z}^s_0$ and $\mathbf{z}^q_0$, with $\mathbf{z}^s_0$ selected using the given support mask. 
    \textbf{2) Prompt Refinement via Gradient Flow}: Using \emph{Mask Decoder}’s logit $d_\phi$, we update the query embeddings by the mask decoder driven gradient flow, resample prompts $P_{t+1}$ from $S_{t+1}$, and iterate for $t{=}0{\ldots}T{-}1$ to form a candidate set of masks $\{\hat{\mathbf{m}}_t\}_{t=1}^{T}$. 
    \textbf{3) Segmentation Mask Selection (Top-1 similarity selection)}: Among the generated masks, we select the one whose mask-aware query embedding $z'^{q}_{t}$ exhibits the highest similarity to the support embedding $z'^{s}$. 
    }
  

  
  \label{fig:overall}
  \vspace{-12pt}
\end{figure*}

This section presents \ourmodel{}, a training-free, test-time refinement framework for promptable segmentation. Starting from support and query embeddings, we run a \emph{mask-decoder–driven gradient flow} that iteratively updates the query embedding and resamples prompts to produce a set of candidate masks; we then apply \emph{Segmentation Mask Selection} via Top-1 support–query similarity to choose the final mask (\cref{fig:overall}).

Below, we introduce the entropy-regularized KL gradient flow, detail the refinement with the segmentation mask decoder and the 
segmentation mask selection, and conclude with the motivation grounded in a convergence analysis.
%
%

\subsection{Gradient flow of entropy-regularized KL-divergences}
\label{sssec:num1}
This subsection briefly describes the construction of the gradient flow of entropy-regularized KL-divergences for mask decoder-driven gradient flow. More details can be found in the overview in \citep{santambrogio2017euclidean} and discriminator gradient flow \citep{ansari2021refining} that forms the basis of our research. 

Let $\mathbf{v}$ be a vector of interest (e.g., a latent vector of a query image), and let $\rho(\mathbf{v})$ and $\mu(\mathbf{v})$ be the candidate and target (ideal) probability density functions (pdf) over $\mathbf{v}$, respectively, assuming that the optimal segmentation mask is distributed over $\mu$ in the latent space. 
For simplicity, we restrict our discussion to pdfs defined on  
$\mathbf{v}$ that belongs to the 2-Wasserstein space $\mathcal{W}_2$. 
 
We now employ the gradient flow to update $\mathbf{v}$ in order to make the candidate distribution $\rho$ approximate the target distribution $\mu$.
Specifically, our focus lies in minimizing the entropy-regularized KL-divergence $\mathrm{F}_{\mu}(\rho)$ between the target distribution  $\mu$ and its candidate distribution $\rho$ as:
\begin{align}
    \label{eqn:obj_func}
    \begin{split}
        \min_{\rho}\mathrm{F}_{\mu}(\rho)
        &= \min_{\rho} \Bigl\{ \mathrm{KL}(\mu \| \rho) - \gamma\,\mathrm{H}(\rho) \Bigr\} \\
        &= \min_{\mathbf{v}} 
        \left\{
            \begin{aligned}
                &\int \log\!\Bigl(\frac{\mu(\mathbf{v})}{\rho(\mathbf{v})}\Bigr) \mu(\mathbf{v})\,d\mathbf{v} \\
                &\quad + \gamma \int \rho(\mathbf{v}) \log\bigl(\rho(\mathbf{v})\bigr)\,d\mathbf{v}
            \end{aligned}
        \right\},                     
    \end{split}
\end{align}
where $\mathrm{KL}(\mu\|\rho)$ represents the KL-divergence between two distributions $\mu$ and $\rho$, $\mathrm{H}(\cdot)$ denotes the entropy function, and $\gamma >0$ is a hyperparmeter that controls the strength of the entropy regularization. Then, the gradient flow of the functional $\mathrm{F}_{\mu}(\rho)$ is given by: 
\begin{align} \label{eqn:grad_flow}
    \begin{split}
         \frac{\partial \rho}{\partial t} = -\nabla \mathrm{F}_{\mu}(\rho_{t}),
    \end{split}
\end{align}
where the subscript $t$ refers to the time (iteration) index throughout this work. Equivalently, \cref{eqn:obj_func} follows a partial differential equation as:
\begin{align} \label{eqn:pde}
    \begin{split}
         \partial_{t} \rho_{t}(\textbf{v}) &- \nabla_{\textbf{v}} \cdot \left( \rho_{t}(\textbf{v}) \nabla_{\textbf{v}} \log( \rho_{t}(\textbf{v})/\mu(\textbf{v})  )  \right) \\ &- \gamma \Delta_{\textbf{v}\textbf{v}}\rho_{t}(\textbf{v}) = 0,
    \end{split}
\end{align}
where $\nabla_{\textbf{v}}$ and $\Delta_{\textbf{v}\textbf{v}}$ denote the divergence and the Laplace operators, respectively \citep{ansari2021refining}. The \cref{eqn:pde}, which is a type of Fokker-Plank equation~\citep{risken1996fokker}, 
can be rewritten as an equivalent stochastic differential equation (SDE) defined as: 
\begin{align} \label{eqn:sde}
    \begin{split}
         d\textbf{v}_{t} = - \nabla_{\textbf{v}} \log( \rho_{t}(\textbf{v})/\mu(\textbf{v}) )dt + \sqrt{2\gamma}d\textbf{w}_t,
    \end{split}
\end{align}
where $\textbf{w}_t$ follows a standard Wiener process \citep{risken1996fokker}. 
 
We can simulate a sample $\textbf{v}_{0} \sim \rho_{0}=\rho$ using  \cref{eqn:sde} to obtain a sample close to $\mu$. In practice, this simulation is approximated using the Euler-Maruyama method~\citep{kloeden1992stochastic}: 
\begin{align} \label{eqn:euler}
    \begin{split}
         \textbf{v}_{t+1} = \textbf{v}_{t} - \eta \nabla_{\textbf{v}} \log(\rho_{t}(\textbf{v})/\mu(\textbf{v})) + \sqrt{2 \gamma \eta} \mathbf{\xi}_{t}, 
    \end{split}
\end{align}
where $\eta > 0$ is the step size, $\mathbf{\xi}_{t} \sim \mathcal{N}(\textbf{0}, \textbf{I})$, and $t \in \left[0,T\right]$ for the predefined number of iterations $T$. Furthermore, we can address intractability of $\mu(\mathbf{v})$ in  \cref{eqn:euler} by approximating the density ratio 
$\frac{\rho_{t}(\textbf{v})}{\mu(\textbf{v})}$ as 
$\frac{\rho_{0} \textbf{v})}{\mu(\textbf{v})}$, as 
$\frac{\rho_{t}(\textbf{v})}{\mu(\textbf{v})} \approx \frac{\rho_{0}(\textbf{v})}{\mu(\textbf{v})}$ 
for small $t$ and $\eta \to 0$. Given a prediction head $D_{\phi}(\textbf{v})$ (e.g., mask decoder parameterized by $\phi$) that represents the conditional probability of $\textbf{v}$ being a sample from $\mu$, the following expression provides an approximation for the density ratio \citep{sugiyama2012density}: 
\begin{align} \label{eqn:density_ratio}
    \begin{split}
         \frac{\rho_{0}(\textbf{v})}{\mu(\textbf{v})}  &= \frac{1-D_{\phi}(\textbf{\textbf{v}})}{D_{\phi}(\textbf{v})} = \exp{(-d_{\phi}(\textbf{v}))}.
    \end{split}
\end{align} 
Substituting  \cref{eqn:density_ratio} into  \cref{eqn:euler} yields the following results:
\begin{align} \label{eqn:final_sde}
    \begin{split}
         \textbf{v}_{t+1} &= \textbf{v}_{t} - \eta \nabla_{\textbf{v}} \log \left (\frac{1-D_{\phi}(\textbf{\textbf{v}})}{D_{\phi}(\textbf{v})} \right ) + \sqrt{2 \gamma \eta} \mathbf{\xi}_{t} \\
         &= \textbf{v}_{t} + \eta \nabla_{\textbf{v}} d_{\phi}(\textbf{v}) + \sqrt{2 \gamma \eta} \mathbf{\xi}_{t},
    \end{split}
\end{align} 
where $d_{\phi}(\textbf{v}) = -\log((1-D_{\phi})/D_{\phi})$ is the logit output of the predictor $D_{\phi}$\footnote{Although the $D_{\phi}$ takes prompts such as bounding boxes and prompt points as inputs in addition to $\textbf v$, we omitted them in \cref{eqn:density_ratio} and \cref{eqn:final_sde} for clarity.}.

In this work, $\rho$ and $\mu$ denote the distributions of query embeddings from the VFM encoder and optimal embeddings for the mask decoder, respectively, with $D_{\phi}$ representing SAM's mask decoder (a pixel-wise classifier). While a dedicated discriminator would be more principled for $D_{\phi}$, we employ this lightweight proxy to enable a training-free framework.

\subsection{Prompt refinement via gradient flow}
\label{sssec:num2}
In this section, we elaborate how gradient flow can be adopted for refining predicted masks. 
We describe the general framework of few-shot segmentation using promptable segmentation models, and then discuss our method for improving mask quality using mask decoder gradient flow. 
As illustrated in \cref{fig:overall}, the first step 
is to extract image embeddings of a support image $\textbf{I}^{s}$ and a query image $\textbf{I}^{q}$ by an image encoder $E_{\theta}$ as
\begin{align} \label{eqn:image_embedding}
    \begin{split}
         \textbf{z}_{0}^{s} = E_{\theta}(\textbf{I}^{s}) \ , \quad
         \textbf{z}_{0}^{q} = E_{\theta}(\textbf{I}^{q}), 
    \end{split}
\end{align} 
where $\textbf{z}_{0}^{s}$ and $\textbf{z}_{0}^{q}$ are initial image embeddings of $\textbf{I}^{s}$ and $\textbf{I}^{q}$, respectively. 
Across the two images, \cref{eqn:image_embedding} yields two embedding sets, $\{\mathbf{z}^{s}_{0,i}\}_{i=1}^{N_s}$ and $\{\mathbf{z}^{q}_{0,j}\}_{j=1}^{N}$, where $N_s$ is the number of pixels within the support mask and $N$ is the total number of pixels in the embedding space. Then, we compute the initial similarity matrix $S_{0}$ between the query embeddings and the support embeddings. The i-${th}$ row and j-${th}$ column of the $S_0$ is computed as: 
\begin{align} \label{eqn:sim}
    \begin{split}
         S_{0}[i,j] = \textit{sim}(\textbf{z}_{0,i}^{s}, \textbf{z}_{0,j}^{q}), 
    \end{split}
\end{align} 

where $sim(\cdot,\cdot)$ denotes 
a similarity function between the two vectors, representing the similarity computation module of the few-shot segmentation model.
Finally, we sample prompts $P_{0}$ representing the position of the target object, and obtain an initially predicted mask $\hat{\textbf{m}}_0$ using a mask decoder $D_{\phi}^{bin}$:
\begin{align} 
    \begin{split}\label{eqn:prompt_sampling}
         P_{0} &= \textit{prompt\_sampler}(S_{0}),
    \end{split}\\
    \begin{split}\label{eqn:mask}
         \hat{\textbf{m}}_{0} &= D_{\phi}^{bin} (\textbf{z}_{0}^{q}; P_{0}), 
    \end{split}
\end{align} 

\begin{figure}[!t]
    \begin{center}        
    
    \begin{subfigure}{0.19\linewidth}
        \centering
        \includegraphics[width=\linewidth]{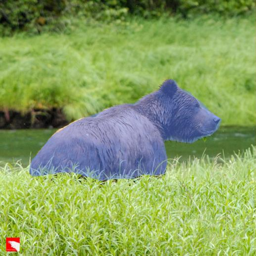}
    \end{subfigure}
    \begin{subfigure}{0.19\linewidth}
        \centering
        \includegraphics[width=\linewidth]{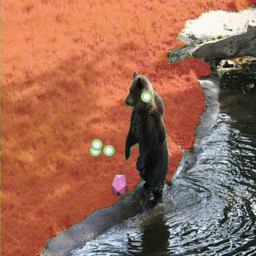}
    \end{subfigure}
    \begin{subfigure}{0.19\linewidth}
        \centering
        \includegraphics[width=\linewidth]{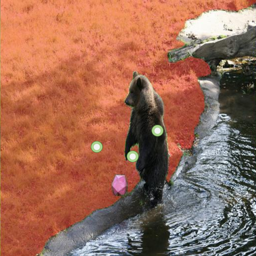}
    \end{subfigure}
    \begin{subfigure}{0.19\linewidth}
        \centering
        \includegraphics[width=\linewidth]{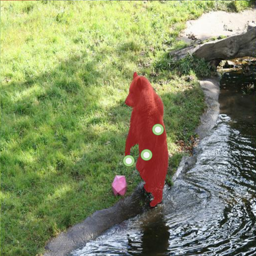}
    \end{subfigure}
    \begin{subfigure}{0.19\linewidth}
        \centering
        \includegraphics[width=\linewidth]{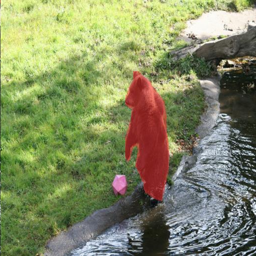}
    \end{subfigure}

    \begin{subfigure}{0.19\linewidth}
        \centering
        \includegraphics[width=\linewidth]{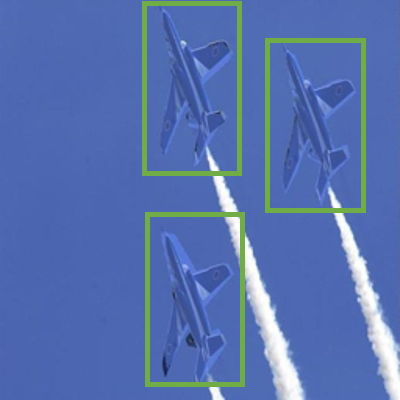}
        \caption{Source}
    \end{subfigure}
    \begin{subfigure}{0.19\linewidth}
        \centering
        \includegraphics[width=\linewidth]{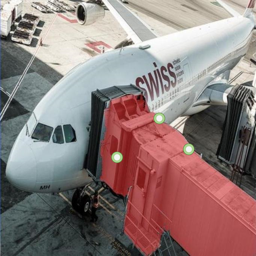}
        \caption{Baseline}
    \end{subfigure}
    \begin{subfigure}{0.19\linewidth}
        \centering
        \includegraphics[width=\linewidth]{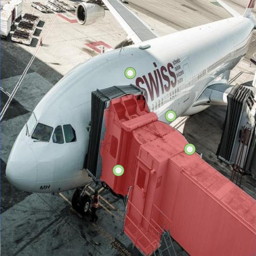}
        \caption{$t=1$}
    \end{subfigure}
    \begin{subfigure}{0.19\linewidth}
        \centering
        \includegraphics[width=\linewidth]{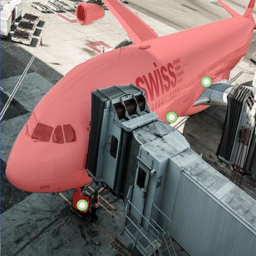}
        \caption{$t=5$}
    \end{subfigure}
    \begin{subfigure}{0.19\linewidth}
        \centering
        \includegraphics[width=\linewidth]{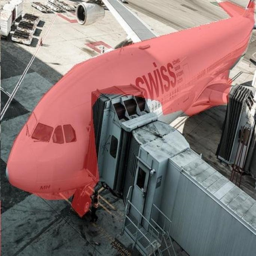}
        \caption{GT}
    \end{subfigure}

    \end{center}
    \vspace{-10pt}
    \caption{{\bf Refinement process of \ourmodel{}.} 
    Curated illustrations of refinement process with PerSAM-F as the baseline. 
    a) Target objects in the support images with blue masks (bear (top) and airplane (bottom)), 
    b) Segmentation results from PerSAM-F with point prompts in green dots. 
    c), d) Refined prompt and segmentation after one and five iterations, 
    e) Ground truth (G.T.).
    }
    \label{fig:ex_refine}
    \vspace{-15pt}
\end{figure}
\noindent where $P_{0}$ is a set of prompts obtained from the similarity map $S_{0}$, and examples of $\textit{prompt\_sampler}$ include top-$k$ sampling \citep{zhang2024personalize} and robust sampler \citep{liu2024matcher}. 
During this process, some prompts in the set $P_{0}$ may be false positives or semantically confusing, and 
several 
methods have been proposed to address this issue, such as cascaded refinement, mask filtering, and robust sampling \citep{zhang2024personalize, liu2024matcher}. 
While these approaches yield reasonably refined masks, 
the results still remain suboptimal (see \cref{fig:intro}). 



Thus, we propose our process for aligning prompts $P_t$ with the query image semantics $z^q_t$ using the gradient flow derived from the mask decoder using \cref{eqn:final_sde}.
In particular, we seek to enhance mask quality by iteratively updating the query embeddings, encouraging that predicted masks are more closely aligned with samples from the true mask distribution. 
Formally, we have the following process in the embedding space:
\begin{align} 
    \begin{split}\label{eqn:mdflow}
         \textbf{z}^{q}_{t+1} &= \textbf{z}^{q}_{t} + \eta \nabla_{\textbf{z}^{q}_{t}} d_{\phi}(\textbf{z}^{q}_{t}, P_{t}) + \sqrt{2 \gamma \eta} \mathbf{\xi}_{t}, 
    \end{split}\\
    \begin{split}\label{eqn:12}
        S_{t+1}[i,j] &= \textit{sim}(\textbf{z}_{0,i}^{s}, \textbf{z}_{t+1,j}^{q}), 
    \end{split}\\
    \begin{split}\label{eqn:13}
        P_{t+1} &= \textit{prompt\_sampler}(S_{t+1}),
    \end{split}\\
    \begin{split}\label{eqn:14}
        \hat{\textbf{m}}_{t+1} &= D_{\phi}^{bin} (\textbf{z}_{0}^{q}; P_{t+1}).
    \end{split}
\end{align}

At the $t$-th iteration, we refine the query embedding $\mathbf{z}_t^q$ via the mask-decoder driven gradient flow in \cref{eqn:mdflow}, update the similarity $S_{t+1}$ as in \cref{eqn:12}, and resample prompts $P_{t+1}$ from $S_{t+1}$ per \cref{eqn:13}. We then decode a candidate segmentation mask $\hat{\mathbf{m}}_{t+1}$ using $D_{\phi}^{bin}$ as in \cref{eqn:14}. Repeating this procedure for $t=0,\ldots,T\!-\!1$ yields a set of candidates $\mathcal{M}=\{\hat{\mathbf{m}}_{t}\}_{t=0}^{T}$. 

Restricting the refinement to the prompt space preserves the generality of our method across different visual-prompt frameworks, where feature-level updates could easily depend on architecture-specific representations. 
Furthermore, the near-optimal assumption of the embedding space does not necessarily hold in practice, and direct modifications of the embedding may introduce instability. 
To achieve a balanced trade-off between robustness and adaptability, 
we use the refined embedding $z^q_t$ only for prompt resampling, 
while decoding is performed with the stable representation $z^q_0$.

\begin{algorithm}[!t]
\caption{\ourmodel{}: Gradient-Flow Prompt Refinement with Top-1 Similarity Selection}
\label{alg:mdflow}
\begin{algorithmic}
\Require image encoder $E_\theta$, mask decoder $D_\phi$, similarity $\textit{sim}$, prompt sampler, iterations $T$
\State \textbf{Init:} Encode support/query $\rightarrow$ $(\mathbf{z}^s_0,\mathbf{z}^q_0)$; compute $S_0$; sample $P_0$.
\State Initialize candidate masks $\mathcal{M}\leftarrow\emptyset$, scores $\mathcal{S}\leftarrow\emptyset$.
\For{$t = 0$ to $T-1$}
  \State \textbf{{Gradient Flow:}} update query embedding (\cref{eqn:mdflow}).
  \State Recompute similarity $S_{t+1}$; resample prompts $P_{t+1}$.
  \State Decode mask $\hat{\mathbf{m}}_{t+1} \leftarrow D^{bin}_\phi(\mathbf{z}^q_0; P_{t+1})$. 
  \State Append $\hat{\mathbf{m}}_{t+1}$ to $\mathcal{M}$.
  \State Score $s_{t+1} \leftarrow \textit{sim}(\text{support},\,\text{query masked by }\hat{\mathbf{m}}_{t+1})$. 
  \State Append $s_{t+1}$ to $\mathcal{S}$.
\EndFor
\State \textbf{Best Mask Selection:} $t^\star \leftarrow \arg\max s_t$; \\
\Return $\hat{\mathbf{m}}^\star \leftarrow \hat{\mathbf{m}}_{t^\star}$
\end{algorithmic}
\end{algorithm}

\subsection{Segmentation mask selection}
\label{ssec:masksel}

For selecting the best segmentation mask, we compute a similarity score $s_t$ between the support embedding and the query embedding aggregated inside $\hat{\mathbf{m}}_{t}$ (e.g., masked average pooling), and choose the final mask by top-1 similarity.

Let $\mathbf{m}^s\in\{0,1\}^N$ be the given support mask and $\hat{\mathbf{m}}_{t}\in\{0,1\}^N$ be the candidate mask at iteration $t$. We obtain embeddings inside a mask, $\mathbf{z}'^s$ and $\mathbf{z}'^q$, using
\begin{align}
    \begin{split}
        \mathbf{z}'^s = E_\theta(\mathbf{I}^s \odot m^s),
    \quad
    \mathbf{z}'^q_t = E_\theta(\mathbf{I}^q \odot \hat{m_t}).         
    \end{split}
\end{align}
We then form the support and query representatives
\begin{align}
\begin{split}
    \bar{\mathbf{z}}'^s \;=\; {{1}\over{|N|}}\sum_{i=1}^{N}\mathbf{z}'^s,  
    \quad
    \bar{\mathbf{z}}'^q_t \;=\; {{1}\over{|N|}}\sum_{i=1}^{N} \mathbf{z}'^q_t.
\end{split}
\end{align}
Given a similarity function $sim(\cdot,\cdot)$, e.g., cosine similarity, we score each candidate by
\begin{align}
\begin{split}
s_t \;=\; sim\!\left(\bar{\mathbf{z}}'^s,\, \bar{\mathbf{z}}'^q_t \right), \quad t=0,\dots,T.
\end{split}
\end{align}
Finally, we select the best iteration $t^*$ via top-1 similarity:
\begin{align}
t^\star \;=\; {\arg \max}_{t\in\{0,\dots,T\}} s_t.
\end{align}
Accordingly, the final mask $\hat{\mathbf{m}}^\star$ is obtained as $\hat{\mathbf{m}}^\star = \hat{\mathbf{m}}_{t^\star}$.
Our overall algorithm is summarized in \cref{alg:mdflow}, and \cref{fig:ex_refine} illustrates the progressive refinement process  
where the points located at the same position are pruned.

\subsection{Motivation for the gradient flow and top-1 similarity selection}
\label{sec:motivation}
Our goal is to refine prompts by updating the query embedding via a gradient flow, and then to select a reliable mask among the candidates generated across iterations. The design is motivated by a convergence perspective.
    
\noindent \textbf{Motivation for gradient-flow updates.} First of all, we demonstrate theoretical convergence of \ourmodel{} in the following proposition, whose proof is given in Appendix A. 
\begin{proposition}
\label{th:1}
Let $\rho_t$ be a candidate pdf in 2-Wasserstein space $\mathcal{W}_2$ being updated according to the gradient flow of the entropy-regularized KL-divergence $F_{\mu}(\rho)$ with local minimum $\mu^{*}$. If the initial pdf $\rho_0$ lies in a neighborhood of $\mu^{*}$, then $\rho_t$ converges to $\mu^{*}$ exponentially as $t \to \infty$.
\end{proposition}


Now let $\rho_t$ denote the distribution of query embeddings at iteration $t$ and let $\mu^{\star}$ be a (local) optimum that is most consistent with the mask decoder. Under the proximity assumption that the initial embedding distribution $\rho_0$ lies in a neighborhood of $\mu^{\star}$ (which is a plausible scenario for a pre-trained VFM), the entropy-regularized KL gradient flow drives $\rho_t$ toward $\mu^{\star}$ at an exponential rate (\cref{th:1}). Hence, even a small number of iterations improves the mask quality, which justifies our use of gradient flow refinement.

\noindent \textbf{Motivation for top-1 similarity selection.}
The proximity assumption does not always hold in practice (\cref{sec:fail}). We empirically observe cases where $\rho_0$ is not sufficiently close to $\mu^{\star}$, leading to suboptimal or unstable refinements. To cope with such violations, we record a set of candidate masks over $T$ iterations and introduce a simple yet effective top-1 similarity selection: among the candidates, we choose the mask whose corresponding query embedding is most similar to the support embedding. This selection step acts as a practical safeguard, retaining gains when the assumption holds while improving robustness when it does not (see sensitivity analysis and convergence failure cases in ~\cref{sec:sensitivity} and ~\cref{sec:fail}, respectively).

\vspace{-5pt} 
\section{Experiments}
\vspace{-5pt} 

We evaluate \ourmodel{} across six benchmark datasets for two separate segmentation tasks. All experiments are conducted under identical protocols to assess segmentation quality and generalization.

\begin{figure*}[!t]
    \centering
    \begin{minipage}[t]{0.48\linewidth}
        \subcaption{Semantic Segmentation\label{fig:semantic}}
        \vspace{0.5em} 
        \centering

        \begin{subfigure}[b]{0.24\linewidth}
            \centering
            \includegraphics[width=\linewidth,height=2.0cm]{figures/main/semantic/PerSAM-f-semantic/ref.png}
        \end{subfigure}
        \begin{subfigure}[b]{0.24\linewidth}
            \centering
            \includegraphics[width=\linewidth,height=2.0cm]{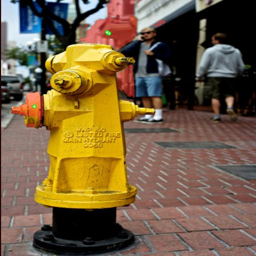}
        \end{subfigure}
        \begin{subfigure}[b]{0.24\linewidth}
            \centering
            \includegraphics[width=\linewidth,height=2.0cm]{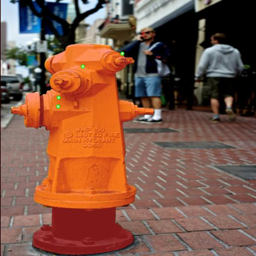}
        \end{subfigure}
        \begin{subfigure}[b]{0.24\linewidth}
            \centering
            \includegraphics[width=\linewidth,height=2.0cm]{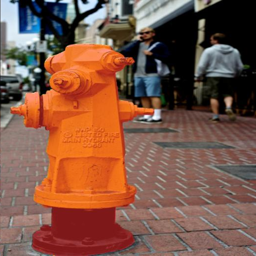}
        \end{subfigure}
        \\[0.1em]

        \begin{subfigure}[b]{0.24\linewidth}
            \centering
            \includegraphics[width=\linewidth,height=2.0cm]{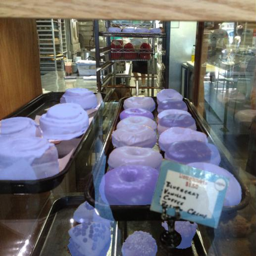}
            \caption*{Support Image}
        \end{subfigure}
        \begin{subfigure}[b]{0.24\linewidth}
            \centering
            \includegraphics[width=\linewidth,height=2.0cm]{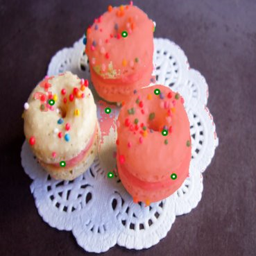}
            \caption*{Baseline}
        \end{subfigure}
        \begin{subfigure}[b]{0.24\linewidth}
            \centering
            \includegraphics[width=\linewidth,height=2.0cm]{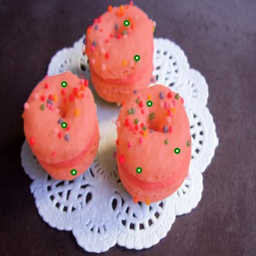}
            \caption*{Refined (Ours)}
        \end{subfigure}
        \begin{subfigure}[b]{0.24\linewidth}
            \centering
            \includegraphics[width=\linewidth,height=2.0cm]{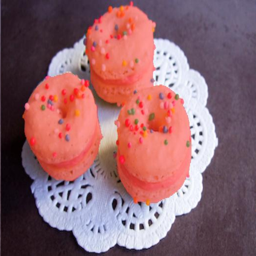}
            \caption*{Ground Truth}
        \end{subfigure}
        \\[0.1em]

    \end{minipage}
    \hspace{1mm}
    \begin{minipage}[t]{0.48\linewidth}
        \subcaption{Part Segmentation\label{fig:part}}
        \vspace{0.5em}
        \centering

        \begin{subfigure}[b]{0.24\linewidth}
            \centering
            \includegraphics[width=\linewidth,height=2.0cm]{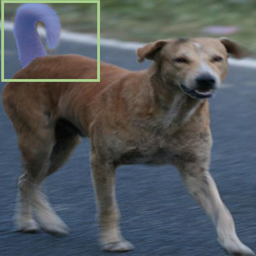}
        \end{subfigure}
        \begin{subfigure}[b]{0.24\linewidth}
            \centering
            \includegraphics[width=\linewidth,height=2.0cm]{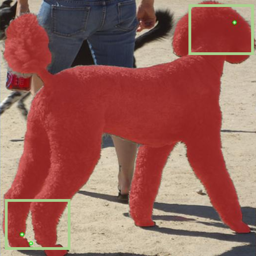}
        \end{subfigure}
        \begin{subfigure}[b]{0.24\linewidth}
            \centering
            \includegraphics[width=\linewidth,height=2.0cm]{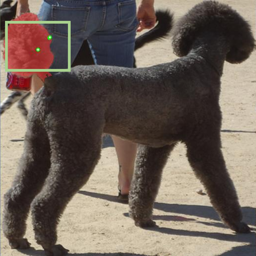}
        \end{subfigure}
        \begin{subfigure}[b]{0.24\linewidth}
            \centering
            \includegraphics[width=\linewidth,height=2.0cm]{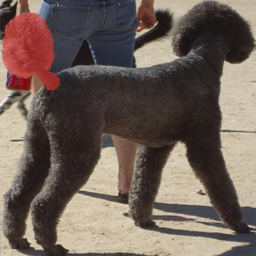}
        \end{subfigure}
        \\[0.1em]

        \begin{subfigure}[b]{0.24\linewidth}
            \centering
            \includegraphics[width=\linewidth,height=2.0cm]{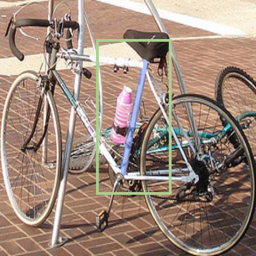}
            \caption*{Support Image}
        \end{subfigure}
        \begin{subfigure}[b]{0.24\linewidth}
            \centering
            \includegraphics[width=\linewidth,height=2.0cm]{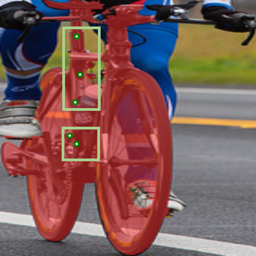}
            \caption*{Baseline}
        \end{subfigure}
        \begin{subfigure}[b]{0.24\linewidth}
            \centering
            \includegraphics[width=\linewidth,height=2.0cm]{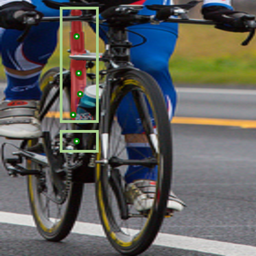}
            \caption*{Refined (Ours)}
        \end{subfigure}
        \begin{subfigure}[b]{0.24\linewidth}
            \centering
            \includegraphics[width=\linewidth,height=2.0cm]{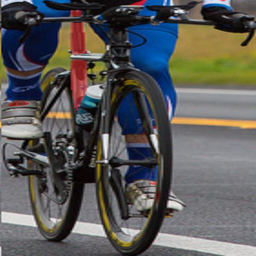}
            \caption*{Ground Truth}
        \end{subfigure}
        \\[0.1em]

    \end{minipage}
    \vspace{-5pt}
    \caption{\textbf{Qualitative results for semantic and part segmentation.} Baselines from top to bottom: PerSAM-F, Matcher.
    \textbf{a)}: Semantic segmentation. 
    Target objects in support images are highlighted with blue masks, and point prompts are denoted by green dots.     
    \textbf{b)}: Part segmentation. Here, target objects are highlighted with blue masks in green boxes, and point prompts are similarly marked and emphasized. 
    Refinement with \ourmodel{} (3$rd$ columns) improves the segmentation over the baselines (2$nd$ columns). 
    }
    \label{fig:combined}
    \vspace{-7pt}
\end{figure*}

\begin{table*}[t]
\centering
\small
\caption{mIoU (\%) on six datasets. B = baseline, O = \ourmodel{} with Oracle selection ($^\ast$), T = \ourmodel{} with Top-1 selection. Bold indicates T values that improve over baseline.}
\vspace{-8pt}
\label{tab:combined_segmentation}
\resizebox{\textwidth}{!}{%
\setlength{\tabcolsep}{4pt}
\begin{tabular}{
l
S[table-format=2.2] S[table-format=2.2] S[table-format=2.2] 
S[table-format=2.2] S[table-format=2.2] S[table-format=2.2] 
S[table-format=2.2] S[table-format=2.2] S[table-format=2.2] 
S[table-format=2.2] S[table-format=2.2] S[table-format=2.2] 
S[table-format=2.2] S[table-format=2.2] S[table-format=2.2] 
S[table-format=2.2] S[table-format=2.2] S[table-format=2.2] 
}
\toprule
& \multicolumn{9}{c}{\textbf{(a) Semantic}} & \multicolumn{9}{c}{\textbf{(b) Part / Fine-grained}} \\
\cmidrule(lr){2-10}\cmidrule(lr){11-19}
& \multicolumn{3}{c}{\textbf{FSS}} & \multicolumn{3}{c}{\textbf{COCO}} & \multicolumn{3}{c}{\textbf{LVIS}}
& \multicolumn{3}{c}{\textbf{PACO}} & \multicolumn{3}{c}{\textbf{Pascal}} & \multicolumn{3}{c}{\textbf{DIS}}\\
\cmidrule(lr){2-4}\cmidrule(lr){5-7}\cmidrule(lr){8-10}
\cmidrule(lr){11-13}\cmidrule(lr){14-16}\cmidrule(lr){17-19}
\multicolumn{1}{c}{\textbf{Method}} &
\multicolumn{1}{c}{\textit{B}} & \multicolumn{1}{c}{\textit{T}} & \multicolumn{1}{c}{\textit{O}$^\ast$} &
\multicolumn{1}{c}{\textit{B}} & \multicolumn{1}{c}{\textit{T}} & \multicolumn{1}{c}{\textit{O}$^\ast$} &
\multicolumn{1}{c}{\textit{B}} & \multicolumn{1}{c}{\textit{T}} & \multicolumn{1}{c}{\textit{O}$^\ast$} &
\multicolumn{1}{c}{\textit{B}} & \multicolumn{1}{c}{\textit{T}} & \multicolumn{1}{c}{\textit{O}$^\ast$} &
\multicolumn{1}{c}{\textit{B}} & \multicolumn{1}{c}{\textit{T}} & \multicolumn{1}{c}{\textit{O}$^\ast$} &
\multicolumn{1}{c}{\textit{B}} & \multicolumn{1}{c}{\textit{T}} & \multicolumn{1}{c}{\textit{O}$^\ast$} \\
\midrule
PerSAM-F     & 85.85 & \textbf{87.26} & 91.88 &
               53.36 & \textbf{55.96} & 63.89 &
               47.62 & \textbf{50.76} & 56.09 &
               45.21 & \textbf{48.17} & 54.20 &
               48.43 & \textbf{51.69} & 57.08 &
               60.90 & \textbf{64.66} & 71.98 \\

Matcher(1-shot) & 92.08 & 92.06 & 93.55 &
              69.53 & \textbf{71.23} & 76.14 &
              59.39 & \textbf{61.52} & 64.75 &
              50.27 & \textbf{54.08} & 56.71 &
              54.76 & \textbf{58.28} & 61.13 &
              46.65 & \textbf{55.08} & 58.10 \\

Matcher(5-shot) & 93.26 & \textbf{93.41} & 94.32 &
              67.69 & \textbf{70.74} & 74.88 &
              57.14 & \textbf{60.79} & 63.88 &
              48.66 & \textbf{53.15} & 55.30 &
              54.54 & \textbf{59.27} & 61.55 &
              \multicolumn{1}{c}{--} & \multicolumn{1}{c}{--} & \multicolumn{1}{c}{--} \\
\bottomrule
\end{tabular}
}
\vspace{-13pt}
\end{table*}

\subsection{Experiment setup}
\label{sec:expset}

\textbf{Datasets.} 
We evaluated \ourmodel{} through two tasks: 
1) semantic segmentation to assess overall object understanding, 
and 2) part segmentation to evaluate detailed context understanding. 
Semantic segmentation performance 
was assessed on three datasets: {\bf FSS-1000} \citep{li2020fss}, {\bf LVIS-92$^{i}$} \citep{liu2024matcher} and {\bf COCO-20$^{i}$} \citep{nguyen2019feature}. 
For part segmentation, we used another three datasets: {\bf PASCAL-Part} \citep{everingham2010pascal,chen2014detect,li2020fss}, {\bf PACO-Part} \citep{liu2024matcher, ramanathan2023paco}, and {\bf DIS5K} \citep{qin2022}. 
We strictly followed the data preprocessing and evaluation protocols outlined in \citep{liu2024matcher}. 


\noindent \textbf{Baselines.}
We utilize PerSAM-F and Matcher as our baseline methods and demonstrate that \ourmodel{} enhances the segmentation mask quality for both methods. 
PerSAM-F uses SAM as its backbone model, fully leveraging it to capture visual clues of target objects in the support images. In contrast, Matcher employs DINOv2 \citep{oquab2023dinov2} with the ViT-L/14 architecture \citep{dosovitskiy2021image} as an image encoder to extract embeddings from both query and support images, and uses SAM's decoder as the segmenter. 

\noindent {\bf Evaluation.} To assess the 
 refinement procedure, we report both top-1 selection and oracle mean Intersection over Union (mIoU) gain of PR-MaGIC over baselines. Oracle results are achieved by selecting the most accurate segmentation mask among $T$$=$$5$ refinement steps. The oracle selection is structurally similar to test-time refinement, providing a meaningful upper bound that highlights the capacity of \ourmodel{}. We provide 
one-shot segmentation performance and additionally provide few-shot (5-shot) results. To further justify our result, we conducted sensitivity analysis on hyperparameters $\eta$ and $T$ in \cref{sec:sensitivity}.



\noindent \textbf{Setup of \ourmodel.} 
Across all the datasets, we set the number of iterations $T=5$. 
The $\gamma$ for entropy regularization was set as 0.1 without any tuning, and 
the $\eta$ for semantic segmentation was 0.001, and that of part segmentation was 0.0001.
These $\eta$ were determined from 10 randomly sampled images 
from left-aside validation sets of COCO-20$^{i}$ and PACO-part. 
The number of points for each method was empirically chosen: 
For semantic segmentation, Matcher used 8 points and PerSAM-F extracted 5 points as prompts. 
For part segmentation, the number of points was set to 5 for the Matcher and to 3 for the PerSAM-F, so that they better localize smaller objects. 
The similarity function $\textit{sim}()$ for \cref{eqn:13} was adopted according to individual baselines \cite{zhang2024personalize,liu2024matcher}. 
Finally, the gradients from \ourmodel{} were clipped to ensure stability throughout the overall process.  
All experiments were conducted with an NVIDIA RTX 6000 ada generation.


\subsection{Experiment results}

We present experiments on semantic and part segmentation, sensitivity analysis with respect to step size and iteration count, and a sample-wise analysis of optimal step variability.
To evaluate the effectiveness of our iterative refinement process, we report three types of results: 
the baseline \textbf{(B)}, \ourmodel{} with Top-1 similarity selection \textbf{(T)}, and \ourmodel{} with Oracle selection \textbf{(O)}. 
The oracle provides the best achievable mIoU across all iterations of \ourmodel{}, serving as an empirical upper bound of our model.
Due to space limitations, the average mIoU progression with a fixed number of iterations is shown in Appendix~B and 1-shot segmentation results across random seeds are provided in Appendix~C.
Furthermore, qualitative results are provided in Appendix~D and E.

\subsubsection{Semantic segmentation}

The semantic segmentation results evaluate how effectively our refinement process enhances object-level segmentation quality across different datasets. 
As summarized in ~\cref{tab:combined_segmentation}, our method consistently improves mIoU on COCO-20$^{i}$, FSS-1000, and LVIS-92$^{i}$. 
The \ourmodel{} improves the baseline by +8.8\%p on FSS, +2.2\%p on COCO, and +2.1\%p on LVIS for PerSAM-F, confirming that the refinement process effectively boosts segmentation accuracy at the object level. 
A small gain observed on FSS-1000 for Matcher(1-shot) is mainly due to its already saturated baseline performance.
In contrast, the refinement yields a pronounced gain for PerSAM-F, showing that our method provides substantial benefits when the initial prediction is less optimal.

Qualitative results in ~\cref{fig:combined}(a) show that baseline predictions often miss object regions or include background noise (e.g., the fire hydrant ($1$st row) or cupcakes ($2$nd row)). 
After refinement, our method captures more precise boundaries and complete object shapes, demonstrating that the proposed refinement process provides consistent improvements without additional training or supervision.
\subsubsection{Part segmentation}

Part segmentation focuses on assessing whether the refinement process can capture fine-grained object structures and precisely delineate relatively small but semantic parts of objects. 
Unlike semantic segmentation, which evaluates holistic object understanding, this task emphasizes local boundaries and subtle structural variations within a single object, which is a more sensitive measure of effectiveness of refinement process with \ourmodel{}.

As shown in the right columns of ~\cref{tab:combined_segmentation}, our method produces consistent and often larger improvements compared to the baseline results. 
The \ourmodel{} improves the baseline by +3.8\%p on PACO and +8.4\%p on DIS for Matcher(1-shot), and by +1.1\%p, +1.2\%p, and +3.2\%p on PACO, Pascal, and DIS for PerSAM-F, respectively. 
These gains indicate improved recovery of fine, part-level regions by our refinement method, especially when baseline predictions lack fine structure or show local fragmentation.

Qualitative results in~\cref{fig:combined}(b) support these observations. When baseline outputs are misaligned or partially miss object parts (e.g., the wheels or handles of a bicycle ($1$st row) and the tail of a poodle ($2$nd row)), the refined predictions yield accurate and continuous part boundaries.
Overall, the stronger relative improvement in part and semantic segmentation suggests that the proposed iterative updates preserve local spatial precision while maintaining global consistency.

\begin{table*}[!tb]
\centering
\small
\caption{Mean $\pm$ std of the optimal iterations in \ourmodel{} under Oracle and Top-1 selection across various datasets and baselines.}
\vspace{-8pt}
\label{tab:mean-variance}
\setlength\tabcolsep{3pt}
\resizebox{\textwidth}{!}{%
\begin{tabular}{l *{6}{cc}}
\toprule
\multirow{2}{*}{\textbf{Model}} &
\multicolumn{2}{c}{\textbf{FSS}} &
\multicolumn{2}{c}{\textbf{COCO}} &
\multicolumn{2}{c}{\textbf{LVIS}} &
\multicolumn{2}{c}{\textbf{PACO-Part}} &
\multicolumn{2}{c}{\textbf{Pascal-Part}} &
\multicolumn{2}{c}{\textbf{DIS}} \\
\cmidrule(lr){2-3} \cmidrule(lr){4-5} \cmidrule(lr){6-7} 
\cmidrule(lr){8-9} \cmidrule(lr){10-11} \cmidrule(lr){12-13}
 & Oracle & Top-1
 & Oracle & Top-1
 & Oracle & Top-1
 & Oracle & Top-1
 & Oracle & Top-1
 & Oracle & Top-1 \\
\midrule
PerSAM 

& 2.53 $\pm$ 1.71 & 2.61 $\pm$ 1.71
& 2.41 $\pm$ 1.72 & 2.51 $\pm$ 1.73
& 2.50 $\pm$ 1.71 & 2.48 $\pm$ 1.71
& 2.50 $\pm$ 1.72 & 2.48 $\pm$ 1.72
& 2.49 $\pm$ 1.72 & 2.48 $\pm$ 1.72
& 2.45 $\pm$ 1.72 & 2.49 $\pm$ 1.69 \\

Matcher (1shot)
 & 2.45 $\pm$ 1.72 & 2.55 $\pm$ 1.71
 & 2.47 $\pm$ 1.74 & 2.47 $\pm$ 1.72
 & 2.05 $\pm$ 1.80 & 2.05 $\pm$ 1.81
 & 2.44 $\pm$ 1.73 & 2.46 $\pm$ 1.73
 & 2.44 $\pm$ 1.73 & 2.42 $\pm$ 1.72
 & 2.51 $\pm$ 1.73 & 2.52 $\pm$ 1.70 \\
Matcher (5shot)
 & 2.50 $\pm$ 1.72 & 2.50 $\pm$ 1.73
 & 2.41 $\pm$ 1.71 & 2.45 $\pm$ 1.74
 & 2.29 $\pm$ 1.77 & 2.29 $\pm$ 1.77
 & 2.49 $\pm$ 1.71 & 2.48 $\pm$ 1.71
 & 2.48 $\pm$ 1.72 & 2.47 $\pm$ 1.72
 & -- & -- \\
\bottomrule
\end{tabular}%
}
\vspace{-5pt}
\end{table*}




\subsection{Sensitivity analysis}
\label{sec:sensitivity}
\begin{figure}[tb]
    \centering
    \vspace{-12pt}
    \includegraphics[width=\linewidth]{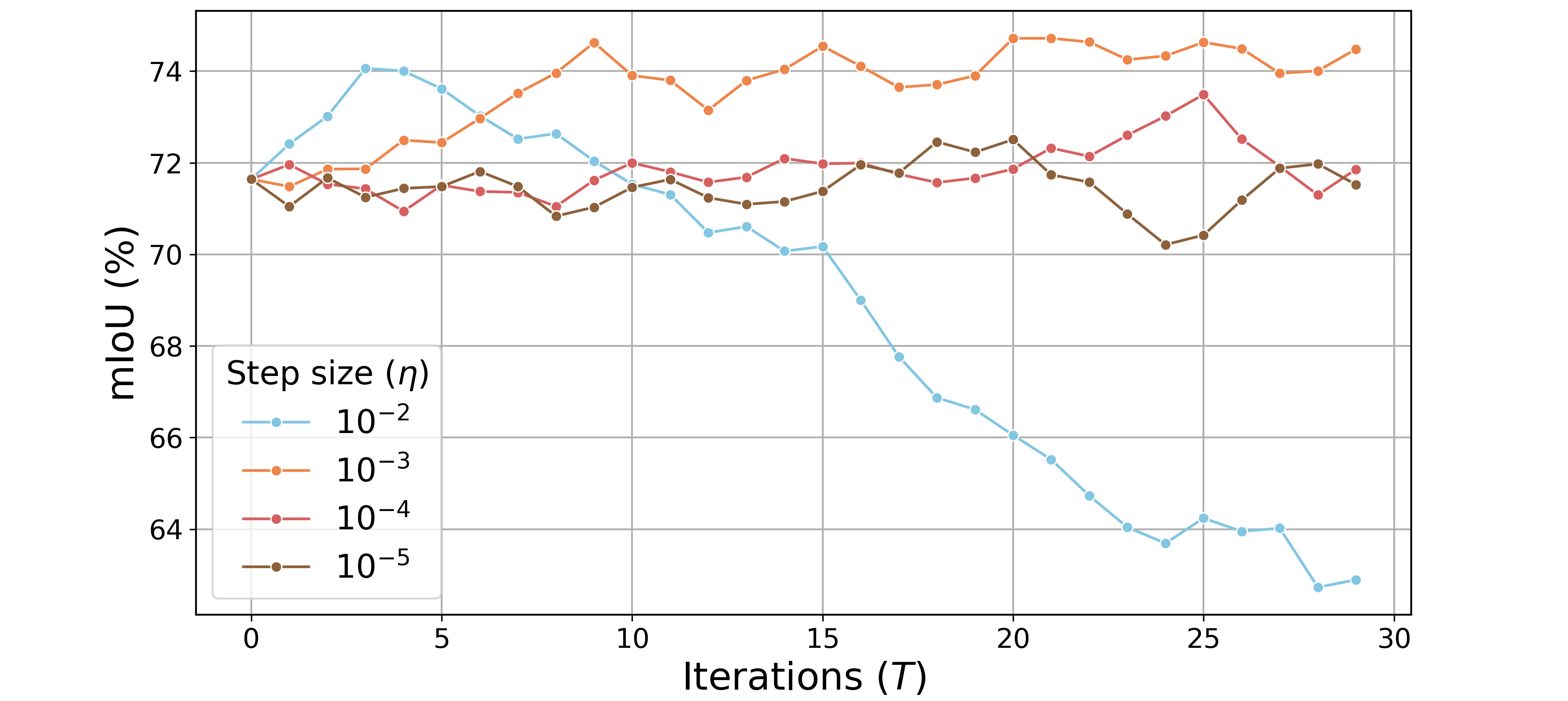}
    \vspace{-15pt}
    \caption{\small {\bf Sensitivity analysis} on the step size $\eta$ and iterations $T$ 
    shows that the gradient flow can substantially improve segmentation performance but may exhibit instability under certain conditions, which motivates top-1 mask selection.}
    \label{fig:sens}
    \vspace{-15pt}
\end{figure}

We randomly sampled 100 examples from the set-aside dataset (COCO-20$^i$ training dataset) to conduct the sensitivity analysis on the step size ($\eta$) and the number of iterations ($T$). For the analysis, we evaluated the mIoU over iterations, testing up to 30 iterations with step sizes of \(10^{-2}\), \(10^{-3}\), \(10^{-4}\), and \(10^{-5}\). \cref{fig:sens} shows a moving average with a window size of 2 to better identify overall trends. 

\noindent \textbf{Findings.} The result demonstrates that the proximity assumption $\rho_0\!\approx\!\mu^{\star}$ is not reliably met in realistic conditions. If the initial embedding distribution $\rho_0$ were consistently close to the decoder-optimal $\mu^{\star}$, performance should improve monotonically with small $T$ (\cref{th:1}). However, our sensitivity results reveal frequent dependency on $\eta$ and $T$, and non-monotonic mIoU trajectories. For instance,
\begin{itemize}
    \item $\eta\!=\!10^{-2}$ yields rapid early gains but degrades with more iterations (instability),
    \item $\eta\!\in\!\{10^{-4},10^{-5}\}$ converges slowly and often saturates below the best achievable mIoU (under-refinement).
\end{itemize}

\noindent \textbf{Implication and design choice.} Given this practical difficulty in convergence, mask selection was introduced as in \cref{ssec:masksel}. 
We run the flow for $T$ iterations, collect the mask from each step as a candidate, and apply top-1 support–query similarity to choose the final output. This selection preserves improvements when refinement helps and mitigates degradation when the flow drifts (see \cref{sec:fail}).


\subsection{Sample-wise optimal step variability and convergence challenges}

We analyze the refinement process by measuring the average iteration (with corresponding standard deviation) at which the optimal performance was achieved (\cref{tab:mean-variance}) for both the oracle and top-1 results.  
The optimal iteration typically occurred around the third refinement step, suggesting that 5 iterations were sufficient to obtain refined outputs. 
However, as shown by the relatively large standard deviations, the optimal iteration varied considerably across samples. 
This variability highlights the necessity of the top-1 selection step, which plays a crucial role in stabilizing and enhancing the overall refinement process.

\label{sec:fail}

\begin{figure}[t]
    \centering
    \begin{subfigure}[b]{0.19\linewidth}
        \centering
        \includegraphics[width=\textwidth, ]{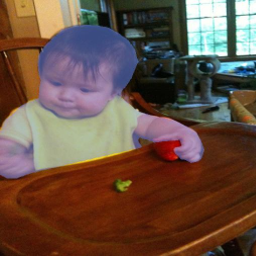}
    \end{subfigure}    
    \begin{subfigure}[b]{0.19\linewidth}
        \centering
        \includegraphics[width=\textwidth, ]{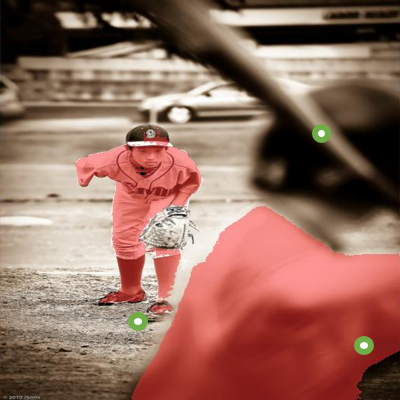}
    \end{subfigure}
    \begin{subfigure}[b]{0.19\linewidth}
        \centering
        \includegraphics[width=\textwidth, ]{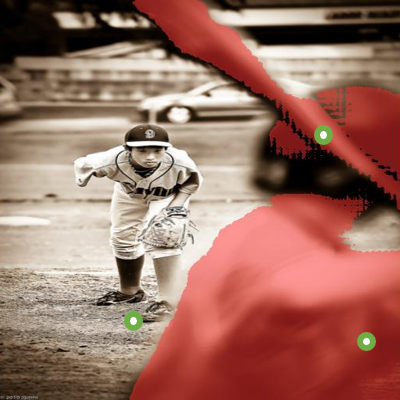}
    \end{subfigure}
    \begin{subfigure}[b]{0.19\linewidth}
        \centering
        \includegraphics[width=\textwidth, ]{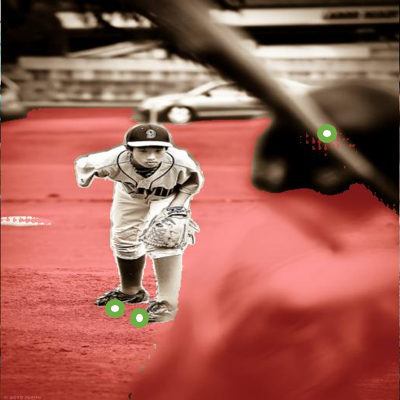}
    \end{subfigure}
    \begin{subfigure}[b]{0.19\linewidth}
        \centering
        \includegraphics[width=\textwidth, ]{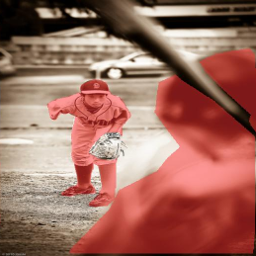}
    \end{subfigure}

    \begin{subfigure}[b]{0.19\linewidth}
        \centering
        \includegraphics[width=\textwidth, ]{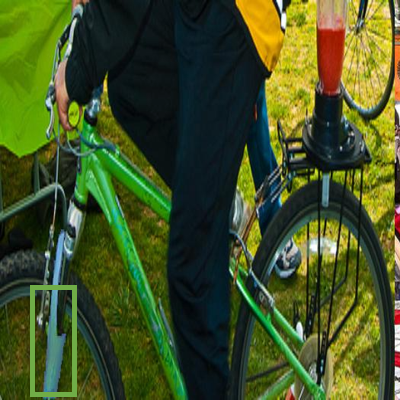}
    \end{subfigure}    
    \begin{subfigure}[b]{0.19\linewidth}
        \centering
        \includegraphics[width=\textwidth, ]{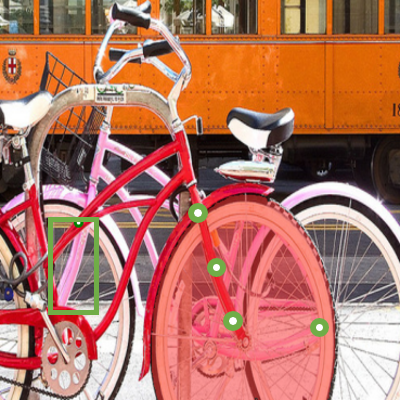}
    \end{subfigure}
    \begin{subfigure}[b]{0.19\linewidth}
        \centering
        \includegraphics[width=\textwidth, ]{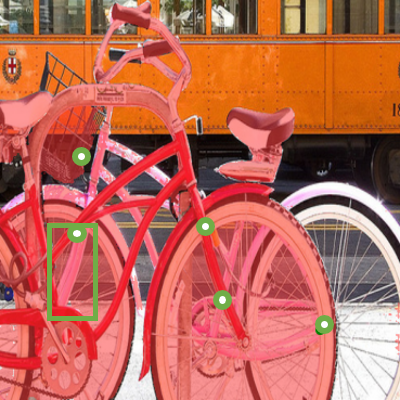}
    \end{subfigure}
    \begin{subfigure}[b]{0.19\linewidth}
        \centering
        \includegraphics[width=\textwidth, ]{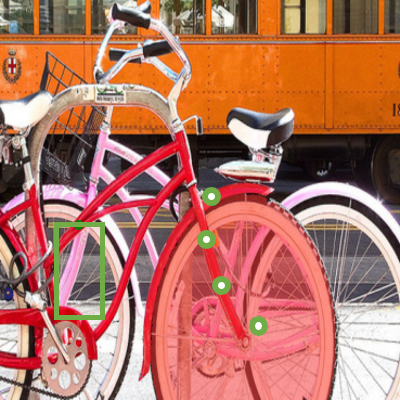}
    \end{subfigure}
    \begin{subfigure}[b]{0.19\linewidth}
        \centering
        \includegraphics[width=\textwidth, ]{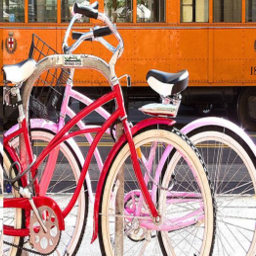}
    \end{subfigure}

    \begin{subfigure}[b]{0.19\linewidth}
        \centering
        \includegraphics[width=\textwidth, ]{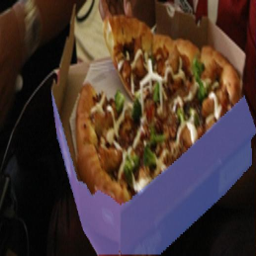}
    \end{subfigure}    
    \begin{subfigure}[b]{0.19\linewidth}
        \centering
        \includegraphics[width=\textwidth, ]{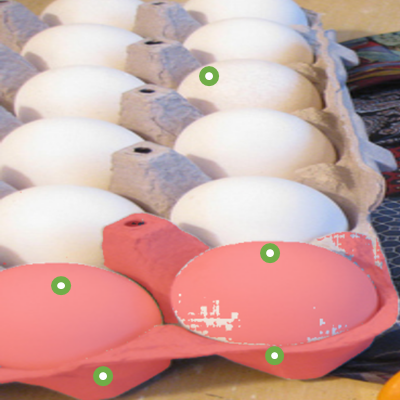}
    \end{subfigure}
    \begin{subfigure}[b]{0.19\linewidth}
        \centering
        \includegraphics[width=\textwidth, ]{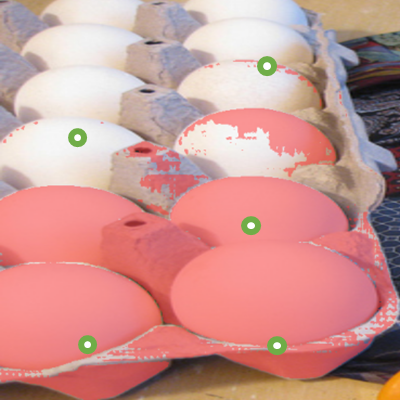}
    \end{subfigure}
    \begin{subfigure}[b]{0.19\linewidth}
        \centering
        \includegraphics[width=\textwidth, ]{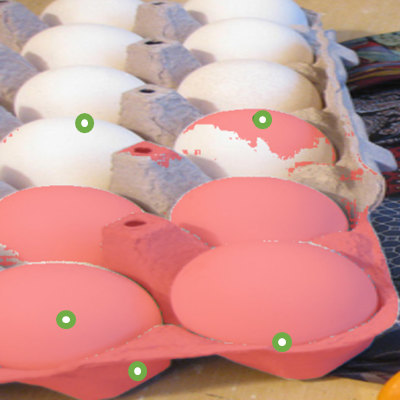}
    \end{subfigure}
    \begin{subfigure}[b]{0.19\linewidth}
        \centering
        \includegraphics[width=\textwidth, ]{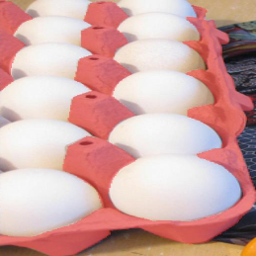}
    \end{subfigure}

    \begin{subfigure}[b]{0.19\linewidth}
        \centering
        \includegraphics[width=\textwidth, ]{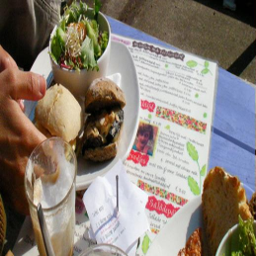}
        \caption{\footnotesize Support}
    \end{subfigure}    
    \begin{subfigure}[b]{0.19\linewidth}
        \centering
        \includegraphics[width=\textwidth, ]{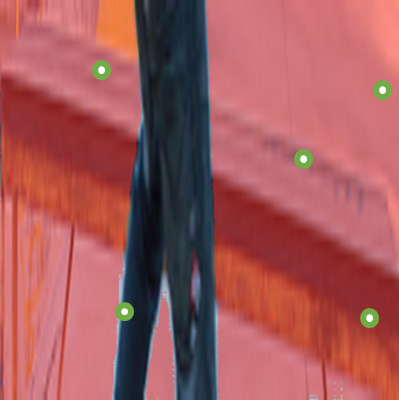}
        \caption{\footnotesize Baseline}
    \end{subfigure}
    \begin{subfigure}[b]{0.19\linewidth}
        \centering
        \includegraphics[width=\textwidth, ]{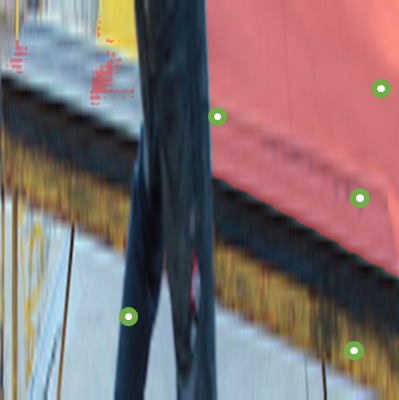}
        \caption{\footnotesize $t=1$}
    \end{subfigure}
    \begin{subfigure}[b]{0.19\linewidth}
        \centering
        \includegraphics[width=\textwidth, ]{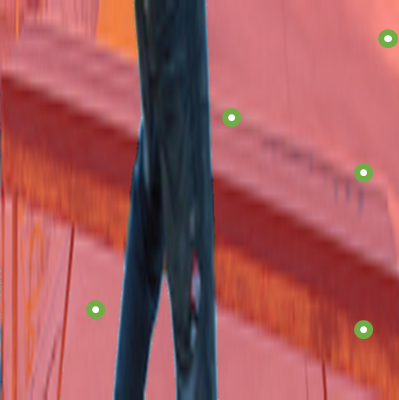}
        \caption{\footnotesize $t=5$}
    \end{subfigure}
    \begin{subfigure}[b]{0.19\linewidth}
        \centering
        \includegraphics[width=\textwidth, ]{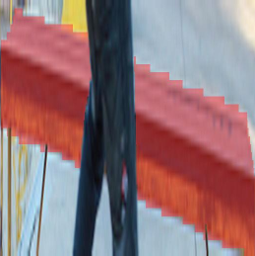}
        \caption{\footnotesize GT}
    \end{subfigure}

    %
    \vspace{-5pt}
\caption{{\bf Convergence failure cases of \ourmodel{}.} Each column represents: (a) Support Image, (b) Baseline output, (c) \ourmodel{} ($t=1$), (d) \ourmodel{} ($t=5$), and (e) Ground Truth. The top two rows compare \ourmodel{} with PerSAM-F as the baseline, while the bottom two rows compare \ourmodel{} with Matcher. In this case, \ourmodel{} selects the top-1 mask from the iterations. } 
\label{fig:fail}
\vspace{-17pt}
\end{figure}

\noindent \textbf{Limitation.} We also highlight challenging scenarios in which \ourmodel{} struggles to converge.
When the visual semantics of the support and query images differ substantially (e.g., the $1$st and $4$th rows of \cref{fig:fail}), prompt refinement becomes particularly difficult and often leads to performance degradation.
Convergence is also fragile when the visual clues in the support image are ambiguous.
As illustrated in \cref{fig:fail}, \ourmodel{} fails to localize fine-grained parts of a bicycle ($2$nd row) and underperforms when the support shows a generic tray while the query target is a specific boundary region ($3$rd row).
These observations collectively emphasize that the top-1 selection serves as a practical safeguard against such convergence failures—preserving the fully automatic advantage of PerSAM and Matcher while enhancing robustness and bridging the gap between theoretical convergence and real-world applicability.

\vspace{-5pt} 
\section{Conclusion}
\vspace{-5pt} 
We present a training-free refinement method that improves segmentation quality using a promptable segmentation model, broadly applicable to auto-prompting frameworks and validated by extensive quantitative and qualitative studies. \ourmodel{} refines the visual clues for SAM’s mask decoder via a gradient-flow update in the query embedding space, enabling better alignment with target semantics. Our analysis shows that, under a proximity assumption, the entropy-regularized KL gradient flow yields rapid improvements with few iterations. 
While sensitivity analyses suggest that our underlying assumption may be fragile in certain cases, the proposed top-1 support–query similarity–based mask selection effectively mitigates this limitation. Importantly, PR-MaGIC remains fully training-free, introducing no learnable parameters, architectural changes, or additional data requirements, yet it achieves consistent improvements in segmentation quality with minimal computational overhead.

\vspace{2pt}
{\small
\noindent\textbf{Acknowledgment.}
This research was supported by RS-2025-02216257 (70\%),
RS-2022-II220290 (20\%), and RS-2019-II191906 (AI Graduate Program at POSTECH, 10\%).
}

{
    \small
    \bibliographystyle{ieeenat_fullname}
    \bibliography{main}
}
\newpage
\appendix
\maketitlesupplementary

\section{Proof of Proposition 1}
\large
\begin{proposition}
\label{th:1}
Let $\rho_t$ be a candidate pdf in 2-Wasserstein space $\mathcal{W}_2$ being updated according to the gradient flow of the entropy-regularized KL-divergence $F_{\mu}(\rho)$ with local minimum $\mu^{*}$. If the initial pdf $\rho_0$ lies in a neighborhood of $\mu^{*}$, then $\rho_t$ converges to $\mu^{*}$ exponentially as $t \to \infty$.
\end{proposition}

\begin{proof}
To prove proposition Proposition 1, We begin by the assumption that $\rho_{0}$ is a neighborhood of $\mu^{*}$, and approximate $F_{\mu}(\rho)$ near $\mu^{*}$ by the second Taylor expansion as: 
\begin{align} \label{eqn:sec_talyor}
    \begin{split}
         F_{\mu}(\rho) &= F_{\mu}(\mu^{*}) + \nabla F_{\mu}(\mu^{*}) (\rho-\mu^{*}) + \frac{1}{2}  \left< F_{\mu}^{''}(\mu^{*})(\rho-\mu^{*}), \rho-\mu^{*} \right> + O \left( \| \rho_t - \mu^* \|^3 \right) \\
                       &= F_{\mu}(\mu^{*}) + \frac{1}{2}  \left< F_{\mu}^{''}(\mu^{*})(\rho-\mu^{*}), \rho-\mu^{*} \right>,
    \end{split}
\end{align}
where $\left< \; , \; \right>$ is an inner product operator, $O$ is the big-O notation, and the second line of \cref{eqn:sec_talyor} holds because $\mu^{*}$ is the minimizer of $F_{\mu}$ (i.e. $\nabla F_{\mu} (\mu^{*})=0$). Furthermore, since $\mu^{*}$ is the minimizer of the functional $F_{\mu}$, the Hessian $F_{\mu}^{''}$ is positive definite. 

Differentiating $F_{\mu}(\rho_t)$ with respective to $\rho_t$ in \cref{eqn:sec_talyor}, we obtain the following approximation for $\nabla F_{\mu}(\rho_{t})$: 
\begin{align} \label{eqn:apporx_grad}
    \begin{split}
         \nabla F_{\mu}(\rho_{t}) \approx F_{\mu}^{''}(\mu^{*}) (\rho_{t}-\mu^{*}).
    \end{split}
\end{align}
The above approximation reduces equation (2) to: 
\begin{align} \label{eqn:apporx_grad_flow}
    \begin{split}
         \frac{\partial \rho_{t}}{\partial t} = -F_{\mu}^{''}(\mu^{*})(\rho_{t}-\mu^{*}).
    \end{split}
\end{align}
To show that \cref{eqn:apporx_grad_flow} describes a linear system with exponential convergence, consider the following differential equation: 
\begin{align} \label{eqn:diff}
    \begin{split}
         \frac{d}{dt} \| \rho_{\tau} -\mu^{*} \|^2 = 2 \left< \rho_{\tau}-\mu^{*}, \frac{\partial \rho_{\tau}}{\partial \tau} \right> 
         = 2 \left< \rho_{\tau}-\mu^{*}, -F_{\mu}^{''}(\mu^{*}) (\rho_{\tau} - \mu^{*}) \right>,
    \end{split}
\end{align}
where $\tau \geq 0$. The second line of the above equation holds by \cref{eqn:apporx_grad}. 
Using the positive definiteness of $F_{\mu}^{''}(\mu^{*})$, the following equation holds for the minimal eigenvalue $\lambda_{min}$ of $F_{\mu}^{''}(\mu^*)$, 
\begin{align} \label{eqn:ineq}
    \begin{split}
         \left< \rho_{\tau}-\mu^{*}, F_{\mu}^{''}(\mu^{*}) (\rho_{\tau} - \mu^{*}) \right> > \lambda_{min} \| \rho_{\tau} - \mu^{*}\|^2.
    \end{split}
\end{align}
Considering \cref{eqn:diff} and \cref{eqn:ineq}, we obtain: 
\begin{align} \label{eqn:dif_eqn}
    \begin{split}
         \frac{d}{dt} \| \rho_{\tau} -\mu^{*} \|^2 < -2 \lambda_{min} \| \rho_{\tau} - \mu^* \|^2.
    \end{split}
\end{align}
By dividing both sides of \cref{eqn:ineq} by $\|\rho_{\tau} - \mu^* \|^2$ and integrating from $\tau=0$ to $\tau = t$, we have:
\begin{align} 
    \begin{split}
        \log \| \rho_t - \mu^* \|^2 < -2 \lambda_{min} t + \log \| \rho_0 - \mu^* \|^2.
    \end{split}
\end{align}
Exponentiating both sides yield:
\begin{align} \label{eqn:dif_eqn_sol}
    \begin{split}
         \| \rho_{t} -\mu^{*}\|^2 < e^{-2 \lambda_{min} t} \| \rho_{0} - \mu^* \|^2,
    \end{split}
\end{align}
which implies that $\rho_{t}$ converges to $\mu^{*}$ exponentially as $t \rightarrow \infty$ when $\rho_0$ is a neighborhood of $\mu^*$. 
\end{proof} 



\newpage
\section{MIoU Progression}




\cref{tab:oracle_top1} reports how mIoU changes across refinement iterations and how these results compare with the Top-1 and Oracle selections. When examining the per-iteration gains, most datasets show only small improvements at fixed iteration counts, and in several cases the refinement even yields slight negative changes. For example, on COCO-20i and LVIS-92i, both Matcher and PerSAM-F frequently exhibit mixed signs across the five iterations, indicating that a fixed number of updates does not reliably improve segmentation quality.

In contrast, Top-1 selection consistently improves upon the baselines across nearly all settings. For instance, Top-1 gains reach +3.05\%p for Matcher 5-shot on COCO-20i, +8.78\%p for PerSAM-F on FSS-1000, and +8.43\%p for Matcher on DIS5K. This improvement comes from selecting the iteration with the highest similarity score for each query–support pair and avoiding the negative effects occasionally introduced by later steps.

The Oracle gains are substantially larger than the Top-1 gains across all datasets, often exceeding +5\%p and reaching up to +14.04\%p for PerSAM-F on FSS-1000. The gap between Top-1 and Oracle stems from the limitation of using support–query similarity as a surrogate for true mask quality. Because this similarity signal can be imperfect, Top-1 sometimes chooses suboptimal iterations even when a better refinement step exists.

Overall, the table indicates three consistent patterns: (1) fixed iteration counts provide inconsistent improvements, (2) Top-1 reliably stabilizes performance across datasets by selecting favorable iterations on a per-sample basis, and (3) Oracle results reveal substantial room for improvement, suggesting that further gains may be achievable with more accurate or robust refinement-selection strategies.

\begin{table}[h]
\centering
\setlength{\tabcolsep}{5pt}
\begin{tabular}{c|l|c|ccccc|cc}
\toprule
\multirow{2}{*}{\textbf{Dataset}} &
\multirow{2}{*}{\textbf{Baselines (\#-shot)}} &
\multicolumn{1}{c|}{\multirow{2}{*}\textbf{Baseline mIoU}} &
\multicolumn{5}{c|}{\textbf{Iterations (\ourmodel{} Gain)}} &
\multirow{2}{*}\textbf{Oracle Gain} & \multirow{2}{*}\textbf{Top-1 Gain} \\ \cmidrule(lr){4-10}
& & & \textbf{1} & \textbf{2} & \textbf{3} & \textbf{4} & \textbf{5} &  &  \\
\midrule\addlinespace[2pt]

\multirow{3}{*}{\textbf{COCO-20i}} 
  & \textbf{Matcher (1shot)}   & \multicolumn{1}{c|}{69.53} & -0.15 & 0.21 & 0.07 & 0.31 & 0.66 & \textbf{+6.61} & \textbf{+1.70} \\
& \textbf{Matcher (5shot)}     & \multicolumn{1}{c|}{67.69} & 0.17  & -0.20 & -0.05 & 0.50 & 0.88 & \textbf{+7.19} & \textbf{+3.05} \\
& \textbf{PerSAM-F (1shot)} & \multicolumn{1}{c|}{53.38} & +0.06 & -0.11 & -0.04 & -0.27 & -0.43 & \textbf{+10.53} & \textbf{+2.60} \\ \hline

\multirow{3}{*}{\textbf{FSS-1000}} 
  & \textbf{Matcher (1shot)}    & \multicolumn{1}{c|}{92.08} & -0.19 & -0.10 & 0.02  & -0.05 & -0.05 & \textbf{+1.47} & \textbf{-0.02} \\
& \textbf{Matcher (5shot)}     & \multicolumn{1}{c|}{93.26} & 0.11  & 0.01  & 0.09  & 0.00  & -0.01 & \textbf{+1.06} & \textbf{+0.15} \\
& \textbf{PerSAM-F (1shot)} & \multicolumn{1}{c|}{85.90} & -0.08 & -0.37 & -0.52 & -0.14 & -0.70 & \textbf{+6.03} & \textbf{+1.41} \\ \hline

\multirow{3}{*}{\textbf{LVIS-92i}} 
  & \textbf{Matcher (1shot)}    & \multicolumn{1}{c|}{59.39} & -0.06 & -0.24 & -0.12 & 0.31 & 0.40 & \textbf{+5.36} & \textbf{+2.13} \\
& \textbf{Matcher (5shot)}     & \multicolumn{1}{c|}{57.14} & -0.05 & -0.04 & 0.17  & 0.54 & 0.97 & \textbf{+6.74} & \textbf{+3.65} \\
& \textbf{PerSAM-F (1shot)} & \multicolumn{1}{c|}{47.59} & +0.03 & +0.03 & -0.15 & -0.19 & -0.11 & \textbf{+8.47} & \textbf{+3.14} \\ \hline

\multirow{3}{*}{\textbf{PACO-Part}} 
  & \textbf{Matcher (1shot)}    & \multicolumn{1}{c|}{50.27} & 0.07  & 0.08  & -0.03 & -0.14 & -0.20 & \textbf{+6.44} & \textbf{+3.81} \\
& \textbf{Matcher (5shot)}     & \multicolumn{1}{c|}{48.66} & 0.07  & 0.08  & -0.03 & -0.14 & -0.20 & \textbf{+6.64} & \textbf{+4.49} \\
& \textbf{PerSAM-F (1shot)} & \multicolumn{1}{c|}{45.27} & -0.04 & +0.01 & -0.09 & -0.12 & +0.01 & \textbf{+8.99} & \textbf{+2.96} \\ \hline

\multirow{3}{*}{\textbf{Pascal-Part}} 
  & \textbf{Matcher (1shot)}    & \multicolumn{1}{c|}{54.76} & 0.05  & 0.00  & 0.30  & 0.04  & -0.04 & \textbf{+6.37} & \textbf{+3.52} \\
& \textbf{Matcher (5shot)}     & \multicolumn{1}{c|}{54.54} & -0.21 & 0.17  & 0.04  & -0.08 & -0.26 & \textbf{+7.01} & \textbf{+4.73} \\
& \textbf{PerSAM-F (1shot)} & \multicolumn{1}{c|}{48.40} & -0.15 & -0.02 & -0.09 & -0.07 & -0.22 & \textbf{+8.65} & \textbf{+3.26} \\ \hline

\multirow{2}{*}{\textbf{DIS5K}} 
  & \textbf{Matcher (1shot)}    & \multicolumn{1}{c|}{46.65} & -0.29 & 0.14  & -0.19 & -0.23 & 0.14  & \textbf{+11.45} & \textbf{+8.43} \\
  & \textbf{PerSAM-F (1shot)} & \multicolumn{1}{c|}{60.32} & +1.06 & +0.55 & +0.21 & -0.24 & +0.23 & \textbf{+11.08} & \textbf{+3.76} \\ \hline

\bottomrule
\end{tabular}
\caption{Average mIoU progression and Oracle/Top-1 improvements across datasets. Gains computed as (Metric - Baseline).}
\label{tab:oracle_top1}
\end{table}







\newpage
\section{One-shot segmentation results across random seeds}
Due to space limitations, we provide additional results across random seeds in the appendix.
\cref{tab:variance_persam} reports the 1-shot segmentation results when PerSAM-F is fixed as the baseline across all evaluated datasets. 
As shown in the Top-1 row, PR-MaGIC consistently improves the mean mIoU over the baseline on every dataset, while maintaining a level of variability comparable to that of the baseline. 
This indicates that the performance gains of PR-MaGIC are achieved in a stable and reliable manner, without introducing substantial seed sensitivity.
\cref{tab:variance-dis} further examines this behavior on the DIS5K dataset using two different baselines, PerSAM-F and Matcher. 
In both cases, PR-MaGIC consistently improves performance over the corresponding baseline, showing that its effectiveness is not tied to a particular baseline method. 
Overall, these results demonstrate that PR-MaGIC provides stable gains across random seeds and generalizes well across different 1-shot segmentation settings.

\begin{table}[!h]
\centering
\vspace{-5pt}
\small
\begin{tabular}{lcccccc}
\hline
 & FSS & COCO & LVIS & PACO-Part & Pascal-Part & DIS5K \\
\hline
Baseline & 85.85 $\pm$ 0.12 & 53.36 $\pm$ 0.14 & 47.62 $\pm$ 0.07 & 45.21 $\pm$ 0.09 & 48.43 $\pm$ 0.05 & 60.90 $\pm$ 0.35 \\
\textbf{Top-1} & $\mathbf{87.26 \pm 0.09}$ & $\mathbf{55.96 \pm 0.25}$ & $\mathbf{50.76 \pm 0.06}$ & $\mathbf{48.17 \pm 0.09}$ & $\mathbf{51.69 \pm 0.06}$ & $\mathbf{64.66 \pm 0.25}$ \\
Oracle & 91.88 $\pm$ 0.07 & 63.89 $\pm$ 0.12 & 56.09 $\pm$ 0.04 & 54.20 $\pm$ 0.02 & 57.08 $\pm$ 0.11 & 71.98 $\pm$ 0.13 \\
\hline
\end{tabular}
\vspace{-5pt}
\caption{Variance of mIoU for PerSAM-F + \ourmodel{} across datasets.}
\label{tab:variance_persam}
\vspace{-10pt}
\end{table}

\begin{table}[!h]
\centering
\vspace{-5pt}
\small
\begin{tabular}{lccc}
\hline
 & Baseline & Top-1 & Oracle \\
\hline
PerSAM-F + \ourmodel{}  &  60.90 $\pm$ 0.35 &  $\mathbf{64.66 \pm 0.25}$ & 71.98 $\pm$ 0.13 \\
Matcher + \ourmodel{} &  47.82 $\pm$ 0.85  & $\mathbf{55.82 \pm 0.64}$  & 59.28 $\pm$ 0.86  \\
\hline
\end{tabular}
\vspace{-5pt}
\caption{Variance of mIoU for DIS5K dataset (1-shot segmentation)}
\label{tab:variance-dis}
\vspace{-10pt}
\end{table}

\newpage
\section{Qualitative results for top-1 selection}
\subsection{Top-1 selection of \ourmodel{} (Matcher)}
\vspace{-10pt}
\begin{figure}[h!]
    \centering
    {
    \setlength{\tabcolsep}{0pt}

    \begin{subfigure}[b]{\textwidth}
        \centering
        \begin{tabular}{@{}cccc@{}}
            \imgbox{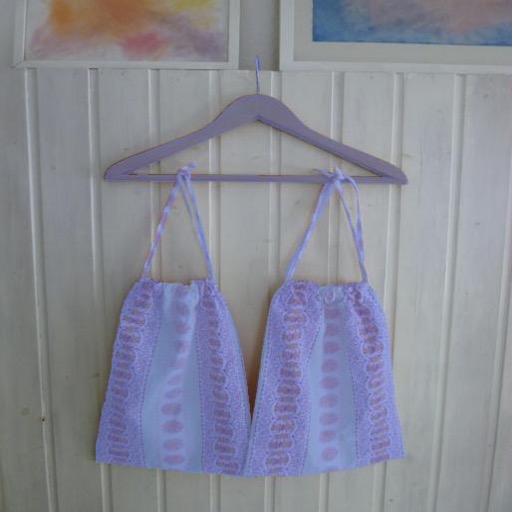} &
            \imgbox{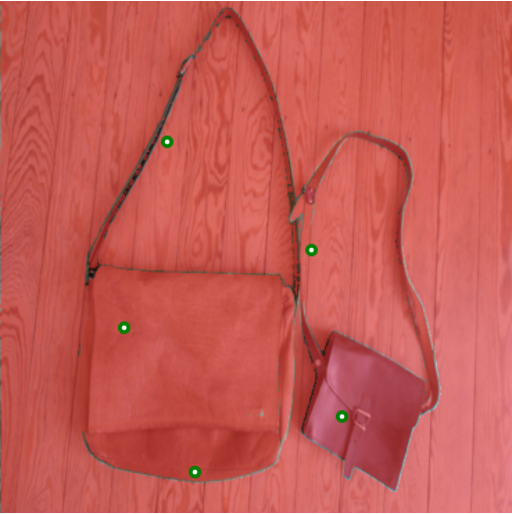} &
            \imgbox{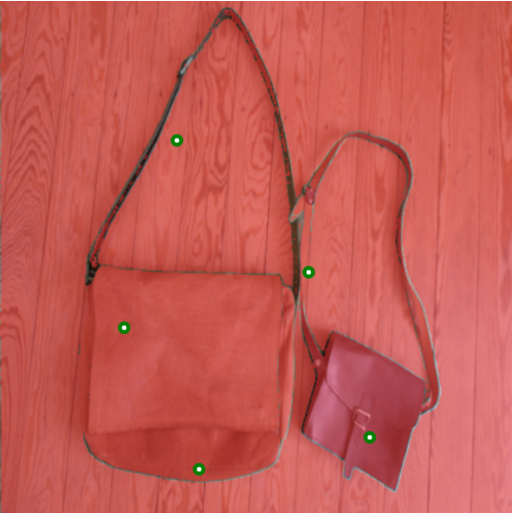} &
            \imgbox{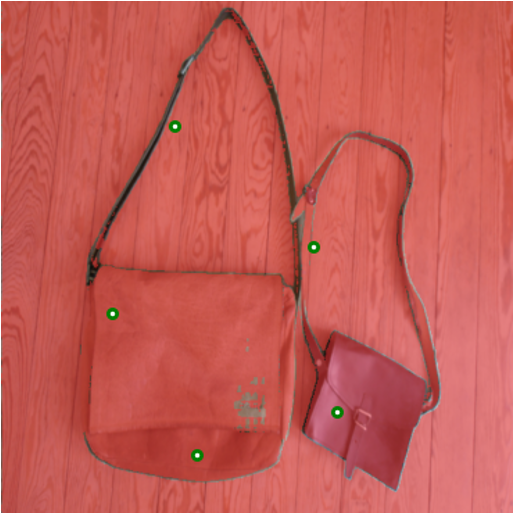} \\
            \scriptsize Support Image & \scriptsize 0 & \scriptsize 1 & \scriptsize 2 \\
            \imgbox{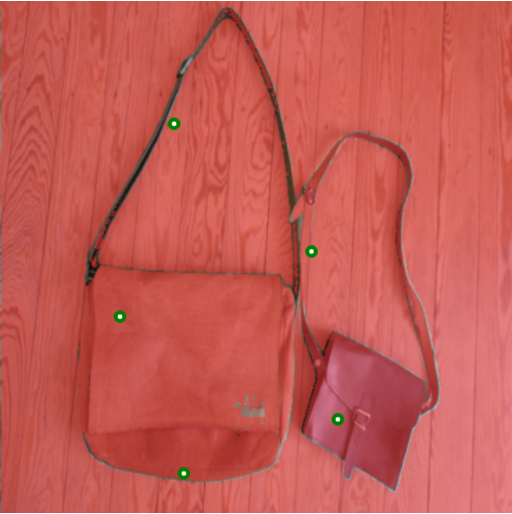} &
            \imgboxsel{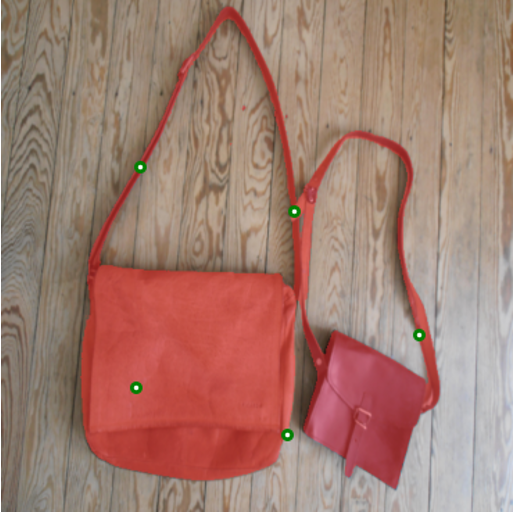} &
            \imgbox{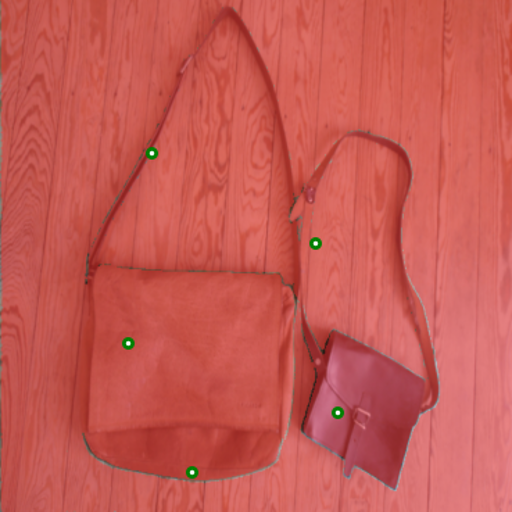} &
            \imgbox{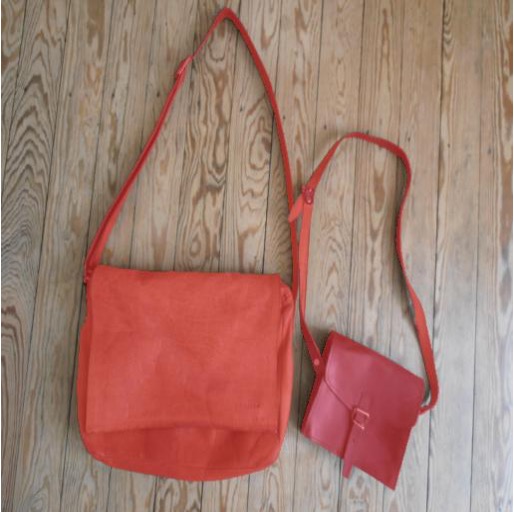} \\
            \scriptsize 3 & \scriptsize 4 (Top-1, Oracle) & \scriptsize 5 & \scriptsize Ground Truth
        \end{tabular}
    \end{subfigure}

    \vspace{0.2em}

    \begin{subfigure}[b]{\textwidth}
        \centering
        \begin{tabular}{@{}cccc@{}}
            \imgbox{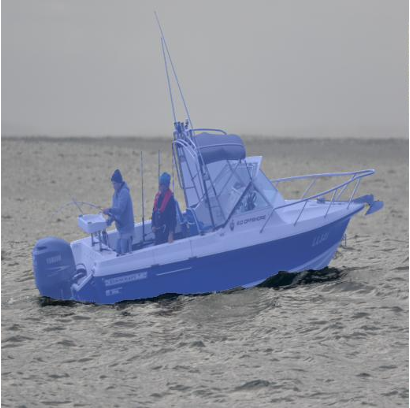} &
            \imgbox{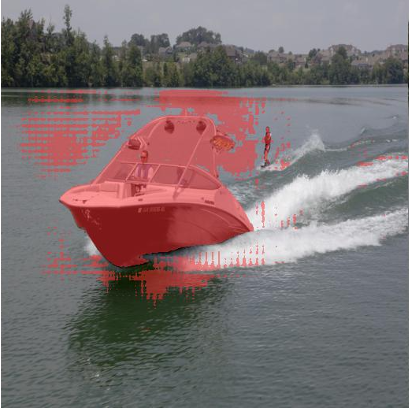} &
            \imgbox{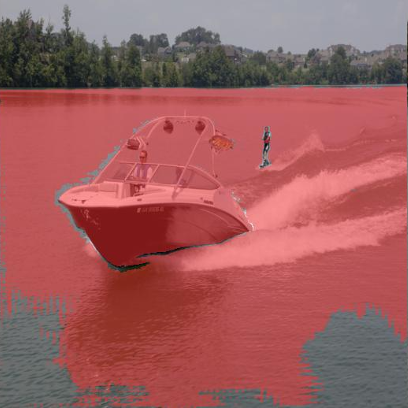} &
            \imgbox{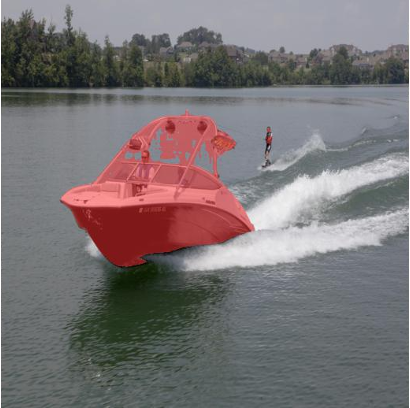} \\
            \scriptsize Support Image & \scriptsize 0 & \scriptsize 1 & \scriptsize 2 \\
            \imgbox{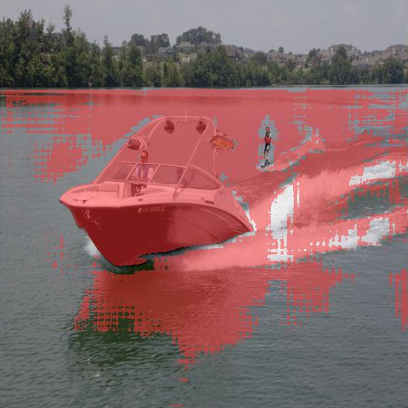} &
            \imgboxsel{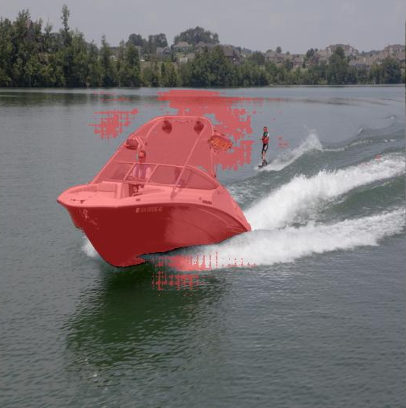} &
            \imgbox{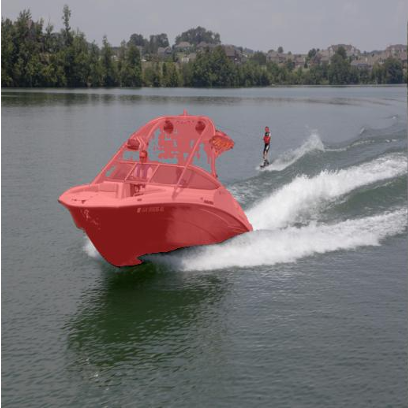} &
            \imgbox{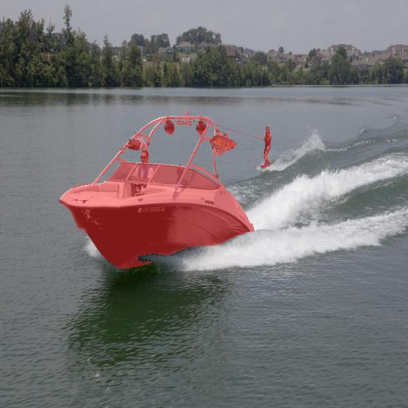} \\
            \scriptsize 3 & \scriptsize 4 (Top-1) & \scriptsize 5 (Oracle) & \scriptsize Ground Truth
        \end{tabular}
    \end{subfigure}

    \vspace{0.2em}

    \begin{subfigure}[b]{\textwidth}
        \centering
        \begin{tabular}{@{}cccc@{}}
            \imgbox{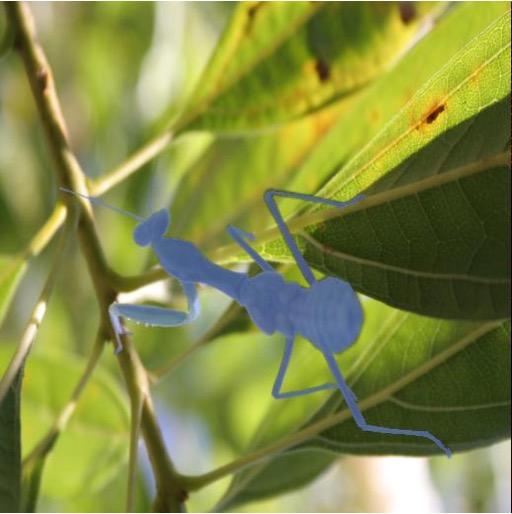} &
            \imgbox{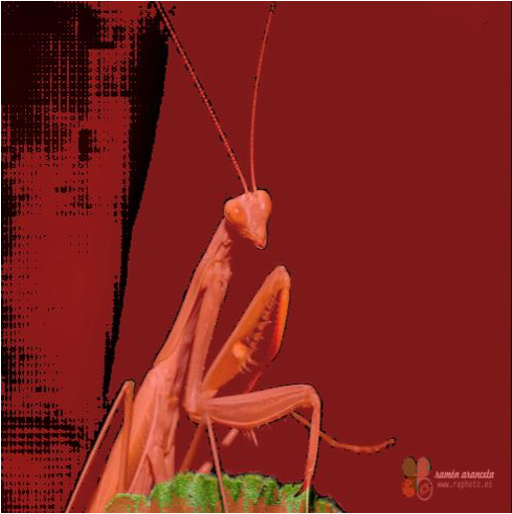} &
            \imgbox{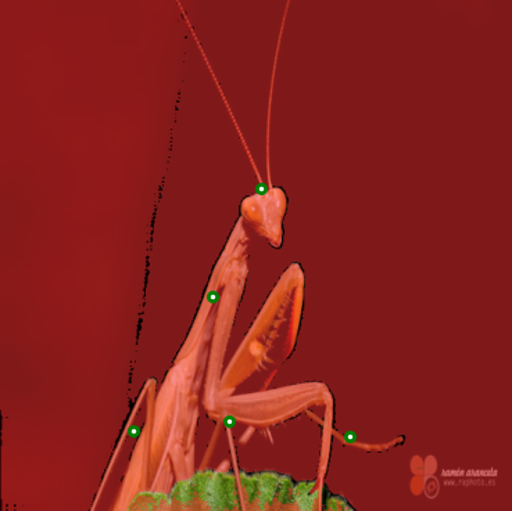} &
            \imgboxsel{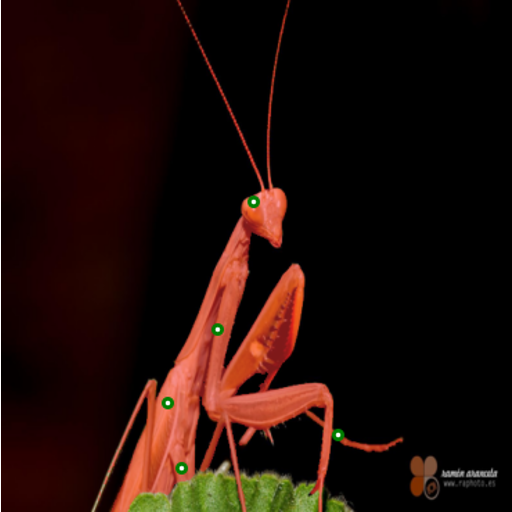} \\
            \scriptsize Support Image & \scriptsize 0 & \scriptsize 1 & \scriptsize 2 (Top-1, Oracle) \\
            \imgbox{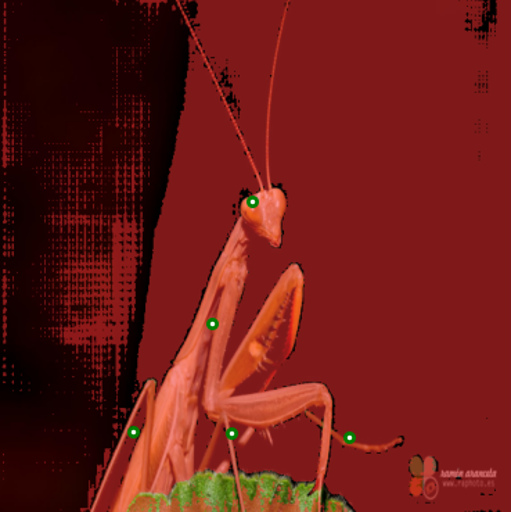} &
            \imgbox{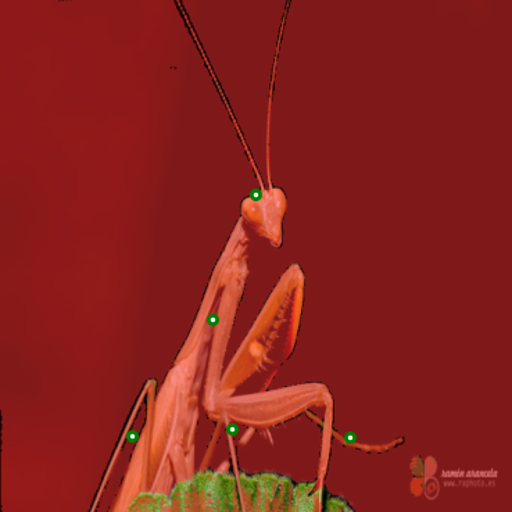} &
            \imgbox{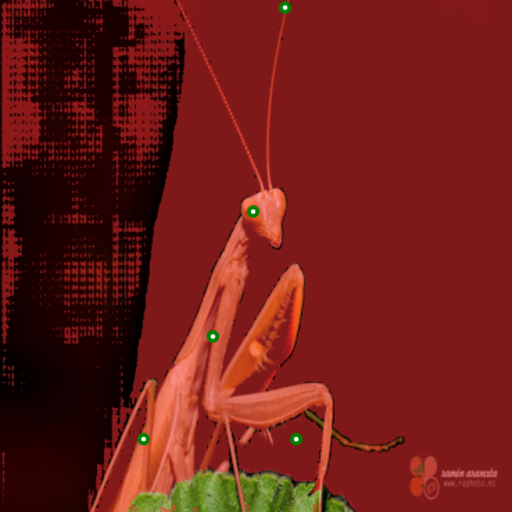} &
            \imgbox{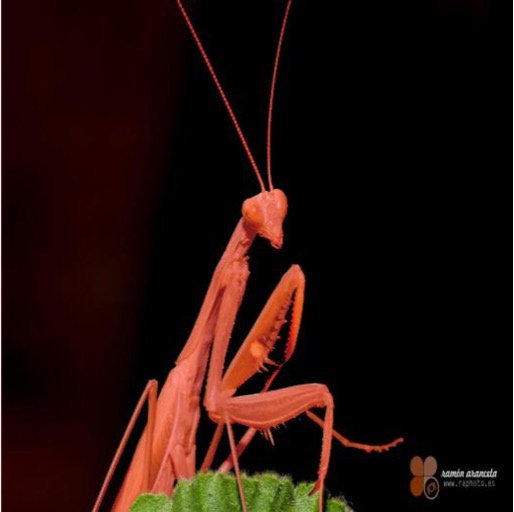} \\
            \scriptsize 3 & \scriptsize 4 & \scriptsize 5 & \scriptsize Ground Truth
        \end{tabular}
    \end{subfigure}

    } 
    \vspace{-10pt}
    \caption{Iterative refinement behavior of \ourmodel{} for Matcher. Each example shows the support image, refinement outputs from iterations 0(baseline) – 5, and the ground truth. The blue-boxed output denotes the step selected by our top-1 selection module. Although the selection is not guaranteed to be globally optimal, it prevents degradation by choosing the final step when refinement converges and an earlier, more stable step when it becomes unstable.}
    \label{fig:matcher_top1_all}
\end{figure}

\newpage
\subsection{Top-1 selection of \ourmodel{} (PerSAM-F)}

\begin{figure}[h!]
    \centering
    {
    \setlength{\tabcolsep}{0pt}

    \begin{subfigure}[b]{\textwidth}
        \centering
        \begin{tabular}{@{}cccc@{}}
            \imgbox{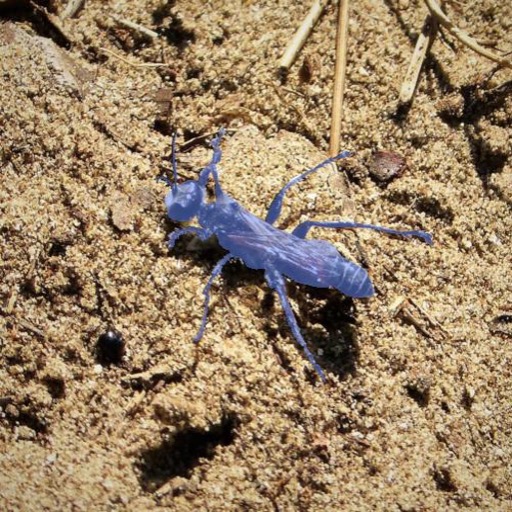} &
            \imgbox{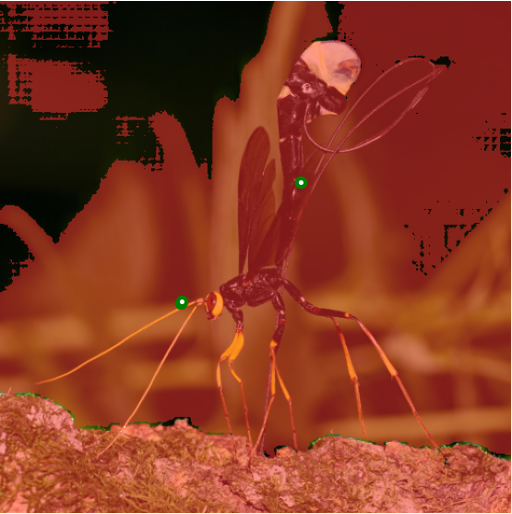} &
            \imgbox{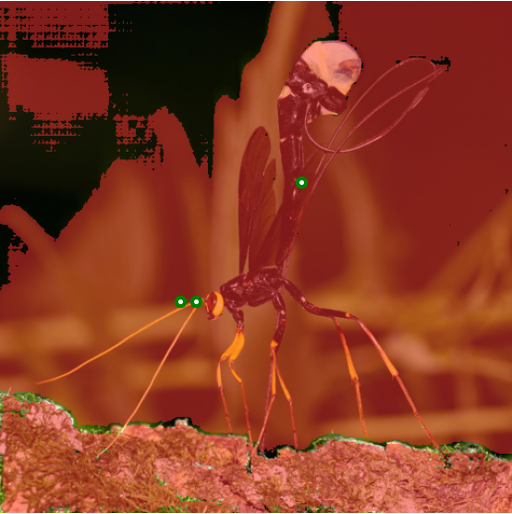} &
            \imgbox{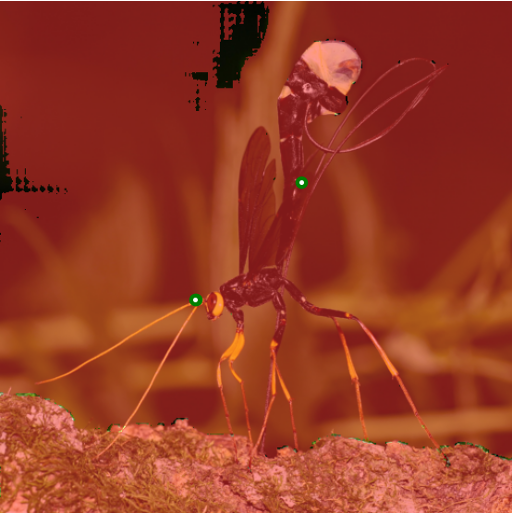} \\
            \scriptsize Support Image & \scriptsize 0 & \scriptsize 1 & \scriptsize 2 \\
            \imgbox{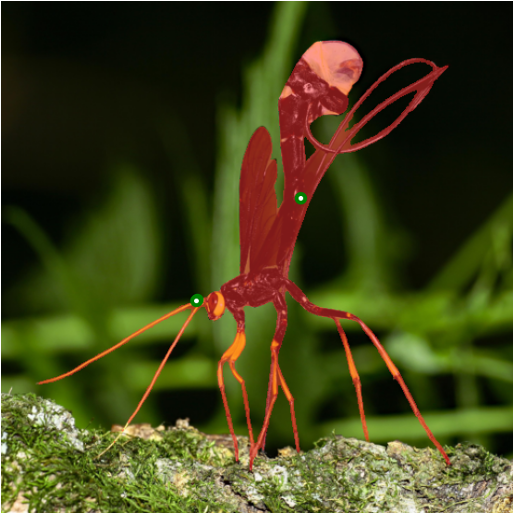} &
            \imgbox{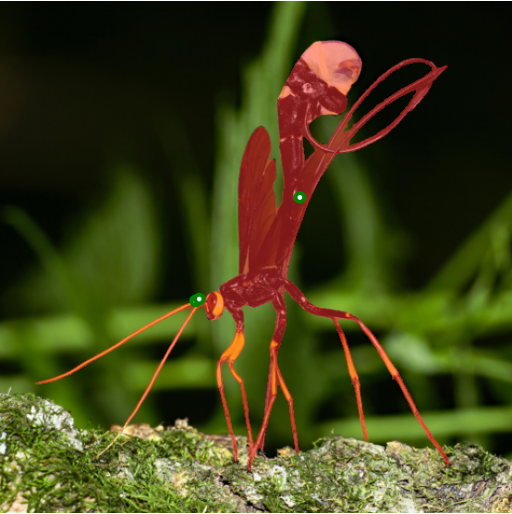} &
            \imgboxsel{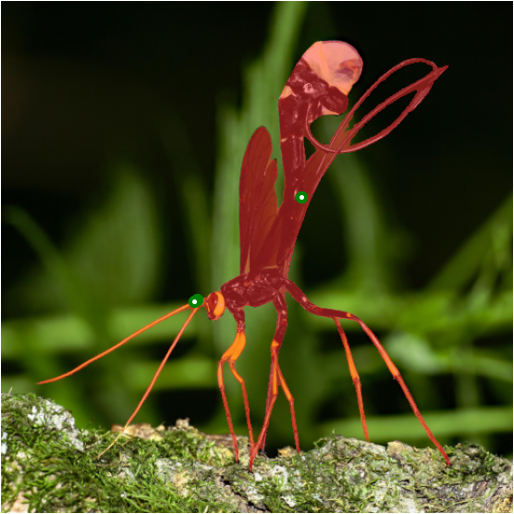} &
            \imgbox{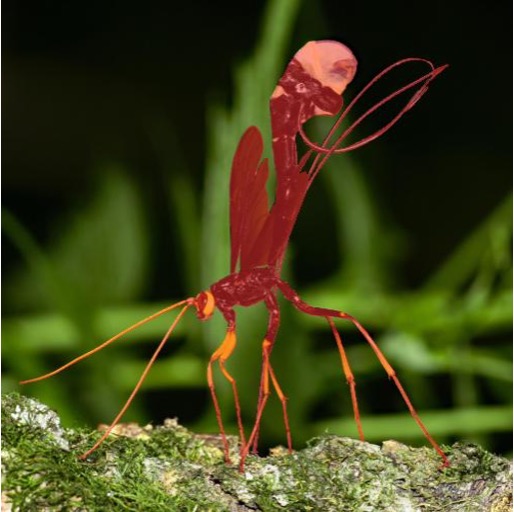} \\
            \scriptsize 3 & \scriptsize 4 & \scriptsize 5 (Top-1, Oracle) & \scriptsize Ground Truth
        \end{tabular}
    \end{subfigure}

    \vspace{0.4em}

    \begin{subfigure}[b]{\textwidth}
        \centering
        \begin{tabular}{@{}cccc@{}}
            \imgbox{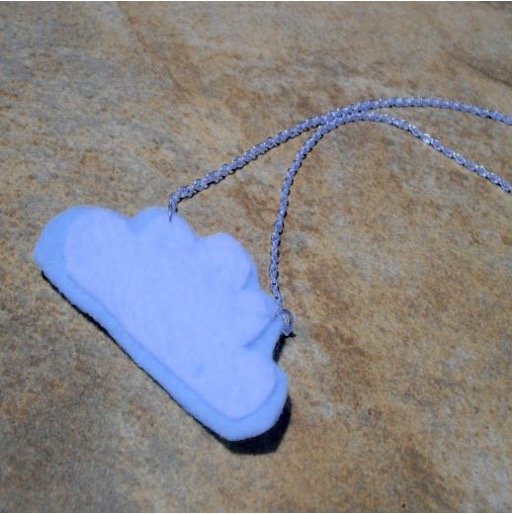} &
            \imgbox{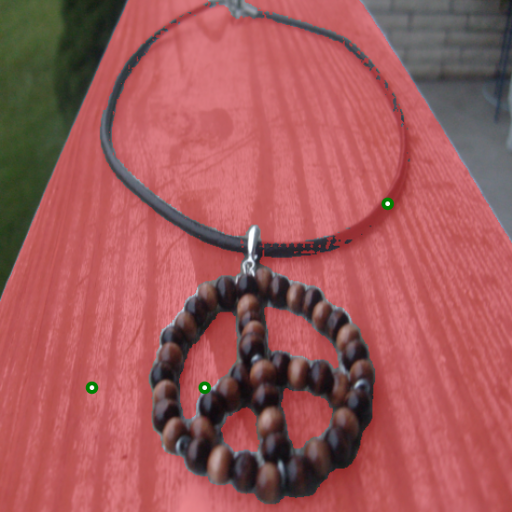} &
            \imgbox{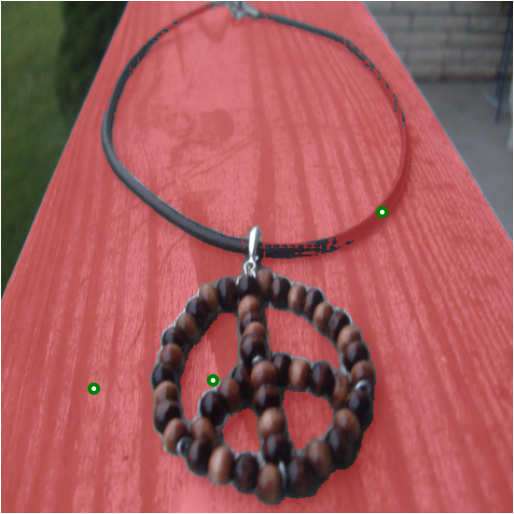} &
            \imgbox{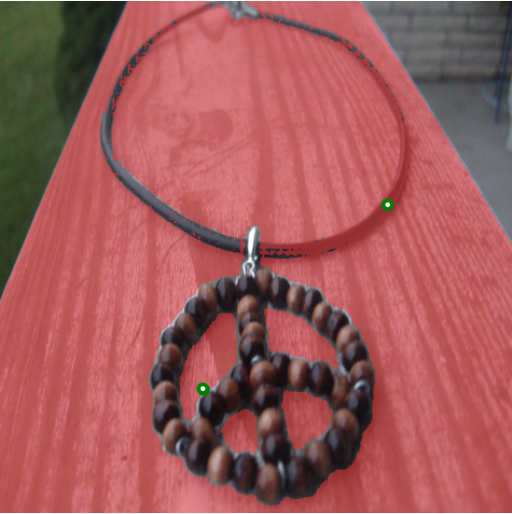} \\
            \scriptsize Support Image & \scriptsize 0 & \scriptsize 1 & \scriptsize 2 \\
            \imgbox{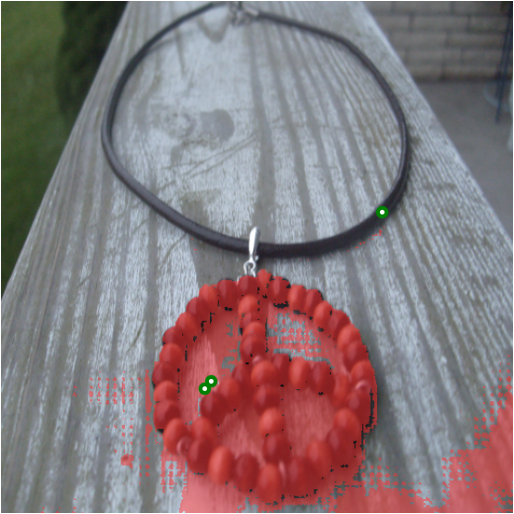} &
            \imgbox{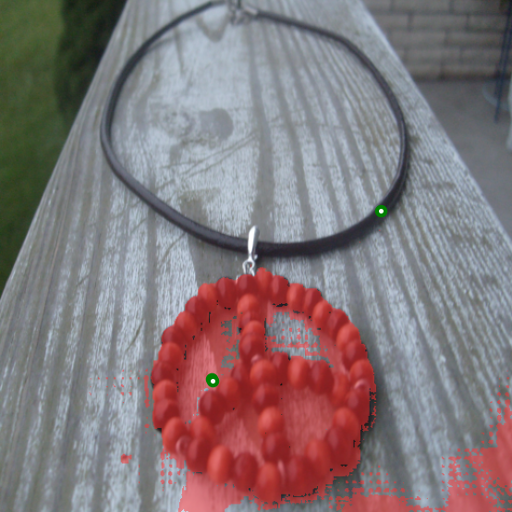} &
            \imgboxsel{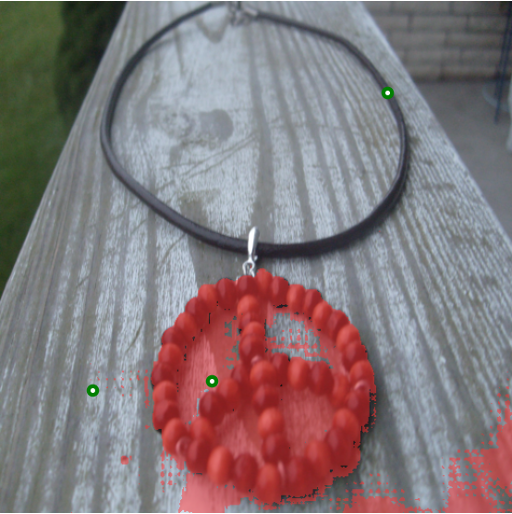} &
            \imgbox{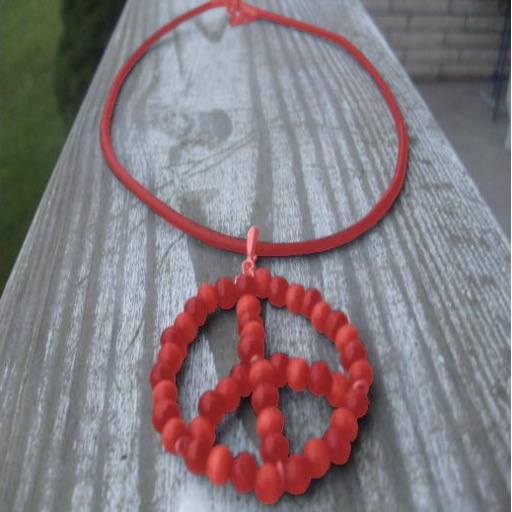} \\
            \scriptsize 3 & \scriptsize 4 & \scriptsize 5 (Top-1, Oracle) & \scriptsize Ground Truth
        \end{tabular}
    \end{subfigure}

    \vspace{0.4em}

    \begin{subfigure}[b]{\textwidth}
        \centering
        \begin{tabular}{@{}cccc@{}}
            \imgbox{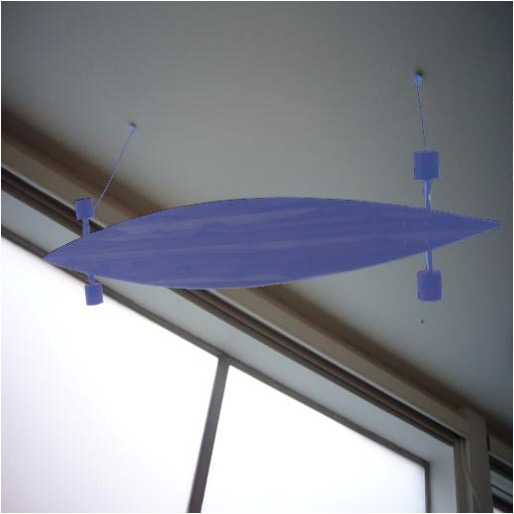} &
            \imgbox{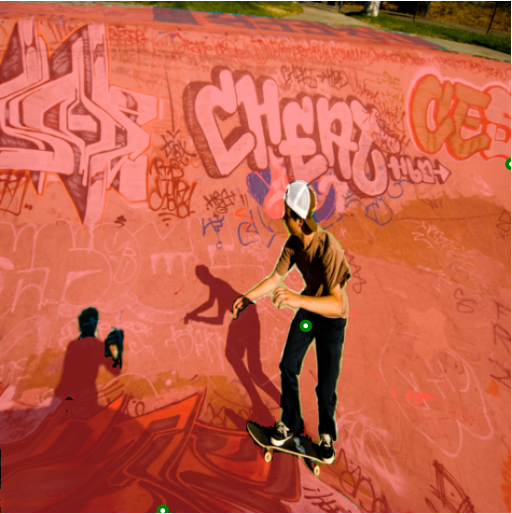} &
            \imgbox{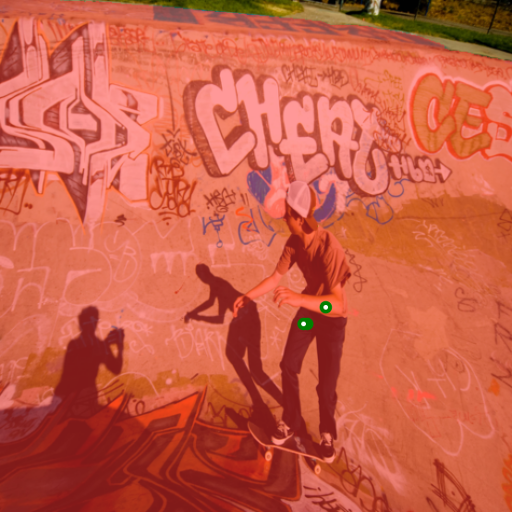} &
            \imgboxsel{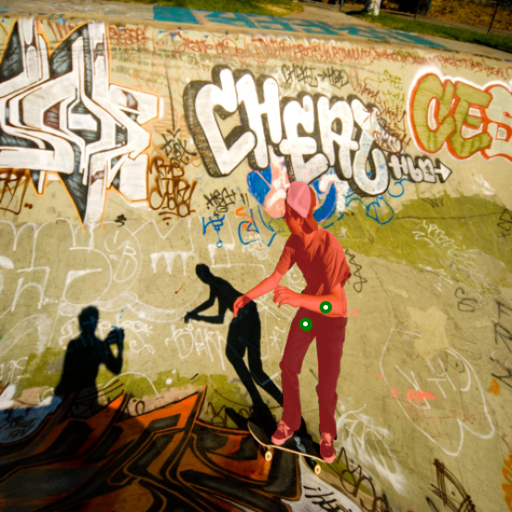} \\
            \scriptsize Support Image & \scriptsize 0 & \scriptsize 1 & \scriptsize 2 (Top-1) \\
            \imgbox{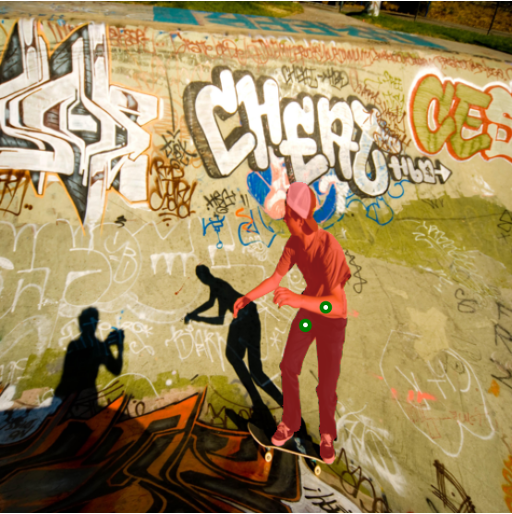} &
            \imgbox{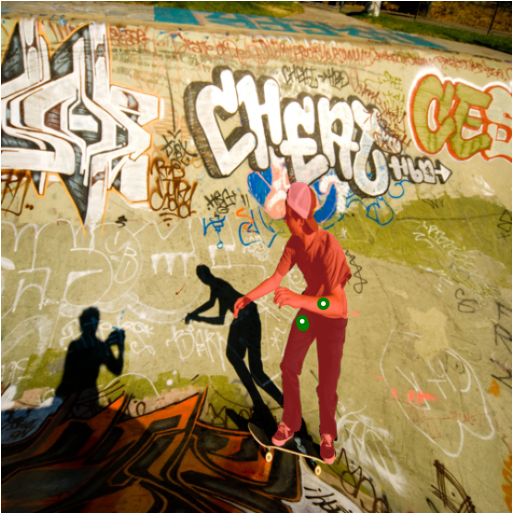} &
            \imgbox{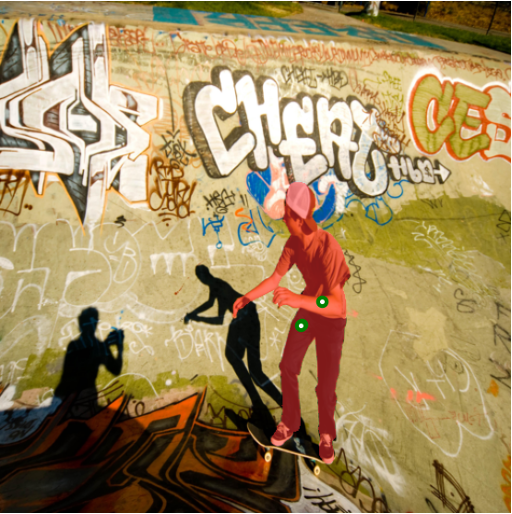} &
            \imgbox{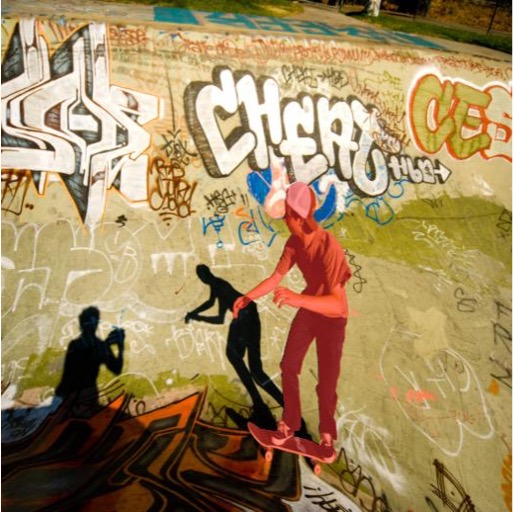} \\
            \scriptsize 3 (Oracle) & \scriptsize 4 & \scriptsize 5 & \scriptsize Ground Truth
        \end{tabular}
    \end{subfigure}

    } 
    \caption{Iterative refinement behavior of \ourmodel{} for PerSAM. Each example shows the support image, refinement outputs from iterations 0(baseline) – 5, and the ground truth. The blue-boxed output denotes the step selected by our top-1 selection module. Although the selection is not guaranteed to be globally optimal, it prevents degradation by choosing the final step when refinement converges and an earlier, more stable step when it becomes unstable.}
    \label{fig:persam_top1_all}
\end{figure}

\vspace{-30pt}

\newpage
\section{Additional Qualitative Results}
\label{sec:appendix:qual}
\vspace{-5pt}

\subsection{One shot semantic segmentation}
\label{sec:sem}
\vspace{-10pt}

\begin{figure}[h!]
    \centering
    
    \begin{subfigure}[b]{0.24\textwidth}
        \centering
        \includegraphics[width=\textwidth, height = 2.8cm]{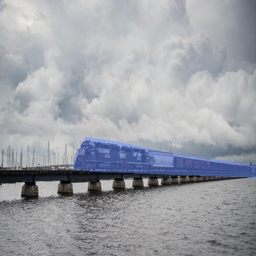}
    \end{subfigure}
    \begin{subfigure}[b]{0.24\textwidth}
        \centering
        \includegraphics[width=\textwidth, height = 2.8cm]{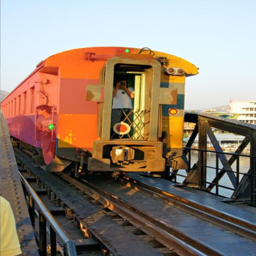}
    \end{subfigure}
    \begin{subfigure}[b]{0.24\textwidth}
        \centering
        \includegraphics[width=\textwidth, height = 2.8cm]{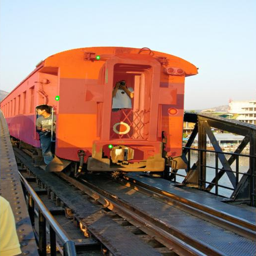}
    \end{subfigure}
    \begin{subfigure}[b]{0.24\textwidth}
        \centering
        \includegraphics[width=\textwidth, height = 2.8cm]{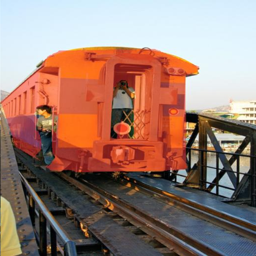}
    \end{subfigure}
    \\
    
     \begin{subfigure}[b]{0.24\textwidth}
        \centering
        \includegraphics[width=\textwidth, height = 2.8cm]{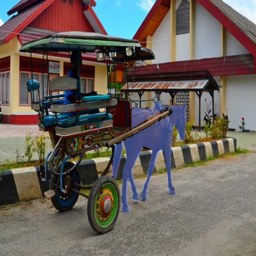}
    \end{subfigure}
    \begin{subfigure}[b]{0.24\textwidth}
        \centering
        \includegraphics[width=\textwidth, height = 2.8cm]{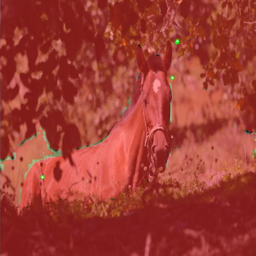}
    \end{subfigure}
    \begin{subfigure}[b]{0.24\textwidth}
        \centering
        \includegraphics[width=\textwidth, height = 2.8cm]{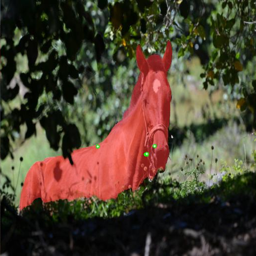}
    \end{subfigure}
    \begin{subfigure}[b]{0.24\textwidth}
        \centering
        \includegraphics[width=\textwidth, height = 2.8cm]{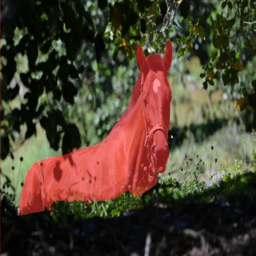}
    \end{subfigure}
    \\
    
    \begin{subfigure}[b]{0.24\textwidth}
        \centering
        \includegraphics[width=\textwidth, height = 2.8cm]{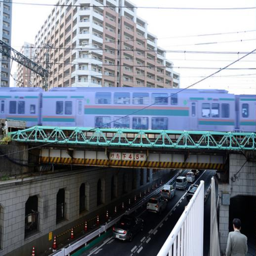}
        \caption{Support Image}
    \end{subfigure}
    \begin{subfigure}[b]{0.24\textwidth}
        \centering
        \includegraphics[width=\textwidth, height = 2.8cm]{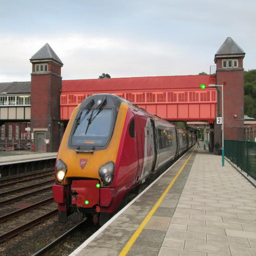}
        \caption{Baseline}
    \end{subfigure}
    \begin{subfigure}[b]{0.24\textwidth}
        \centering
        \includegraphics[width=\textwidth, height = 2.8cm]{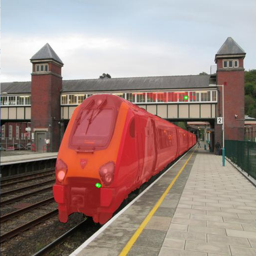}
        \caption{Refined (Ours)}    
    \end{subfigure}
    \begin{subfigure}[b]{0.24\textwidth}
        \centering
        \includegraphics[width=\textwidth, height = 2.8cm]{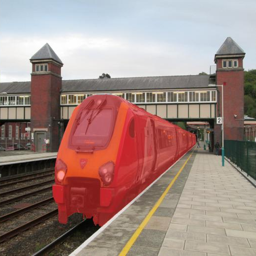}
        \caption{Ground Truth}        
    \end{subfigure}
    \\
    \caption{Qualitative result of \ourmodel{} with PerSAM-F in semantic segmentation.}
\end{figure}

\begin{figure}[h!]
    \centering
     \begin{subfigure}[b]{0.24\textwidth}
        \centering
        \includegraphics[width=\textwidth, height = 3.2cm]{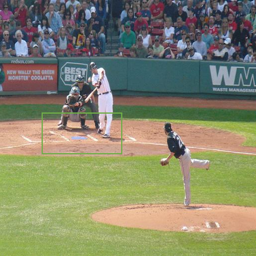}
    \end{subfigure}
    \begin{subfigure}[b]{0.24\textwidth}
        \centering
        \includegraphics[width=\textwidth, height = 3.2cm]{figures/appendix/semantic/1shot-matcher/lvis1_1/before.png}
    \end{subfigure}
    \begin{subfigure}[b]{0.24\textwidth}
        \centering
        \includegraphics[width=\textwidth, height = 3.2cm]{figures/appendix/semantic/1shot-matcher/lvis1_1/after.png}
    \end{subfigure}
    \begin{subfigure}[b]{0.24\textwidth}
        \centering
        \includegraphics[width=\textwidth, height = 3.2cm]{figures/appendix/semantic/1shot-matcher/lvis1_1/gt.png}
    \end{subfigure}
    \\
    
     \begin{subfigure}[b]{0.24\textwidth}
        \centering
        \includegraphics[width=\textwidth, height = 3.2cm]{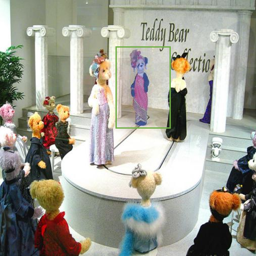}
    \end{subfigure}
    \begin{subfigure}[b]{0.24\textwidth}
        \centering
        \includegraphics[width=\textwidth, height = 3.2cm]{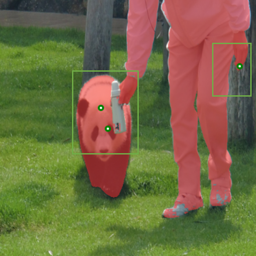}
    \end{subfigure}
    \begin{subfigure}[b]{0.24\textwidth}
        \centering
        \includegraphics[width=\textwidth, height = 3.2cm]{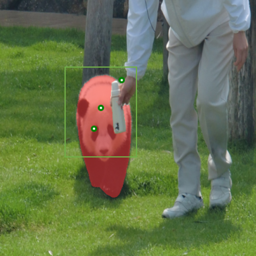}
    \end{subfigure}
    \begin{subfigure}[b]{0.24\textwidth}
        \centering
        \includegraphics[width=\textwidth, height = 3.2cm]{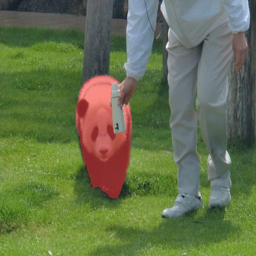}
    \end{subfigure}
    \\
    
     \begin{subfigure}[b]{0.24\textwidth}
        \centering
        \includegraphics[width=\textwidth, height = 3.2cm]{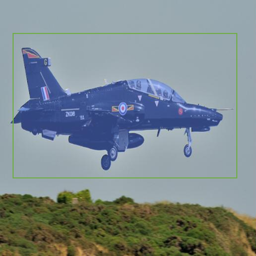}
        \caption{Support Image}
    \end{subfigure}
    \begin{subfigure}[b]{0.24\textwidth}
        \centering
        \includegraphics[width=\textwidth, height = 3.2cm]{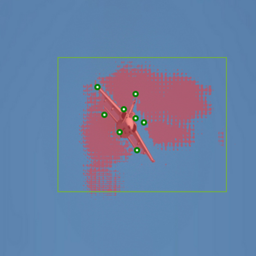}
        \caption{Baseline}
    \end{subfigure}
    \begin{subfigure}[b]{0.24\textwidth}
        \centering
        \includegraphics[width=\textwidth, height = 3.2cm]{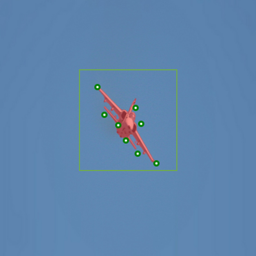}
        \caption{Refined (Ours)}       
    \end{subfigure}
    \begin{subfigure}[b]{0.24\textwidth}
        \centering
        \includegraphics[width=\textwidth, height = 3.2cm]{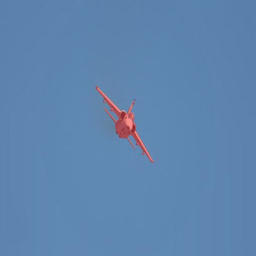}
        \caption{Ground Truth}       
    \end{subfigure}
    \\
   
    \caption{Qualitative result of \ourmodel{} with Matcher in semantic segmentation.}
        \vspace{-15pt}
\end{figure}


\newpage
\subsection{One shot part segmentation}
\label{sec:part}
\vspace{-5pt}

\begin{figure}[h!]
    \centering
    
    \begin{subfigure}[b]{0.24\textwidth}
        \centering
        \includegraphics[width=\textwidth, height = 3.2cm]{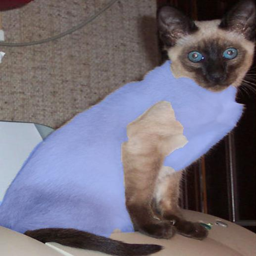}
    \end{subfigure}
    \begin{subfigure}[b]{0.24\textwidth}
        \centering
        \includegraphics[width=\textwidth, height = 3.2cm]{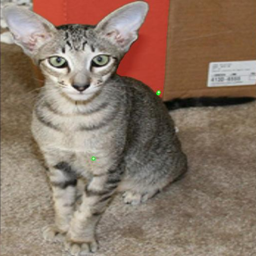}
    \end{subfigure}
    \begin{subfigure}[b]{0.24\textwidth}
        \centering
        \includegraphics[width=\textwidth, height = 3.2cm]{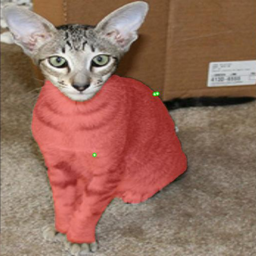}
    \end{subfigure}
    \begin{subfigure}[b]{0.24\textwidth}
        \centering
        \includegraphics[width=\textwidth, height = 3.2cm]{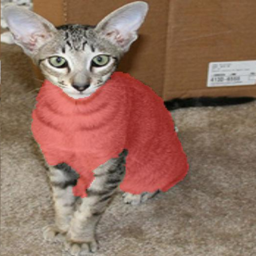}
    \end{subfigure}
    \\
    
    \begin{subfigure}[b]{0.24\textwidth}
        \centering
        \includegraphics[width=\textwidth, height = 3.2cm]{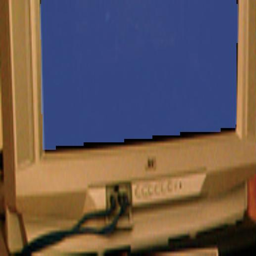}
    \end{subfigure}
    \begin{subfigure}[b]{0.24\textwidth}
        \centering
        \includegraphics[width=\textwidth, height = 3.2cm]{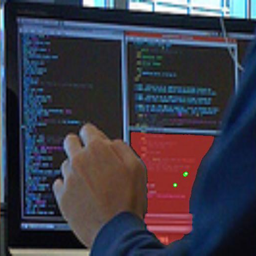}
    \end{subfigure}
    \begin{subfigure}[b]{0.24\textwidth}
        \centering
        \includegraphics[width=\textwidth, height = 3.2cm]{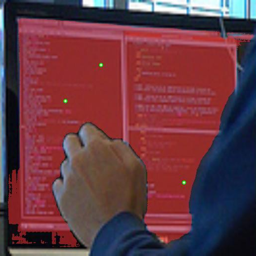}
    \end{subfigure}
    \begin{subfigure}[b]{0.24\textwidth}
        \centering
        \includegraphics[width=\textwidth, height = 3.2cm]{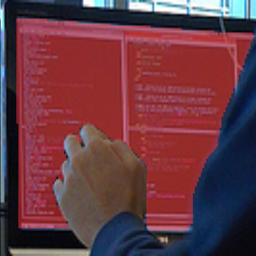}
    \end{subfigure}
    \\
    
    \begin{subfigure}[b]{0.24\textwidth}
        \centering
        \includegraphics[width=\textwidth, height = 3.2cm]{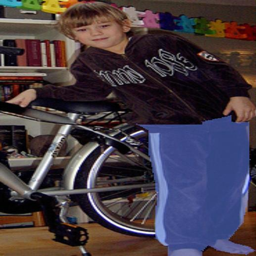}
        \caption{Support Image}
    \end{subfigure}
    \begin{subfigure}[b]{0.24\textwidth}
        \centering
        \includegraphics[width=\textwidth, height = 3.2cm]{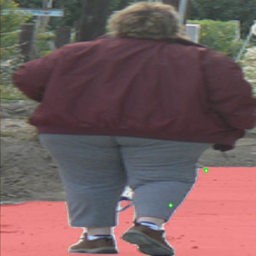}
        \caption{Baseline}
    \end{subfigure}
    \begin{subfigure}[b]{0.24\textwidth}
        \centering
        \includegraphics[width=\textwidth, height = 3.2cm]{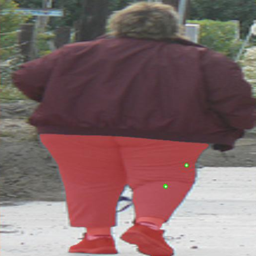}
        \caption{Refined (Ours)}    
    \end{subfigure}
    \begin{subfigure}[b]{0.24\textwidth}
        \centering
        \includegraphics[width=\textwidth, height = 3.2cm]{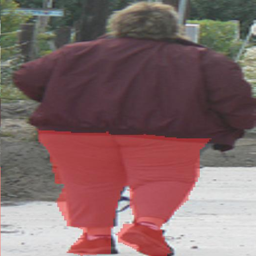}
        \caption{Ground Truth}        
    \end{subfigure}
    \caption{Qualitative result of \ourmodel{} with PerSAM-F in part segmentation.}
\end{figure}

\begin{figure}[h!]
    \centering
    
    \begin{subfigure}[b]{0.24\textwidth}
        \centering
        \includegraphics[width=\textwidth, height = 2.8cm]{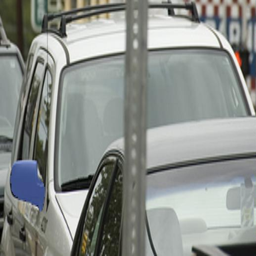}
    \end{subfigure}
    \begin{subfigure}[b]{0.24\textwidth}
        \centering
        \includegraphics[width=\textwidth, height = 2.8cm]{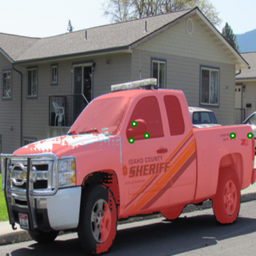}
    \end{subfigure}
    \begin{subfigure}[b]{0.24\textwidth}
        \centering
        \includegraphics[width=\textwidth, height = 2.8cm]{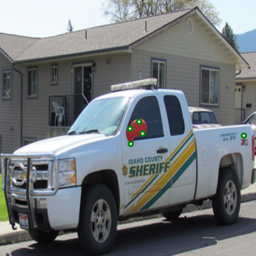}
    \end{subfigure}
    \begin{subfigure}[b]{0.24\textwidth}
        \centering
        \includegraphics[width=\textwidth, height = 2.8cm]{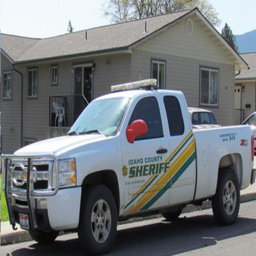}
    \end{subfigure}
    \\
    
    \begin{subfigure}[b]{0.24\textwidth}
        \centering
        \includegraphics[width=\textwidth, height = 2.8cm]{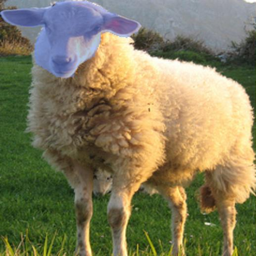}
    \end{subfigure}
    \begin{subfigure}[b]{0.24\textwidth}
        \centering
        \includegraphics[width=\textwidth, height = 2.8cm]{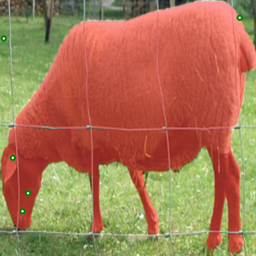}
    \end{subfigure}
    \begin{subfigure}[b]{0.24\textwidth}
        \centering
        \includegraphics[width=\textwidth, height = 2.8cm]{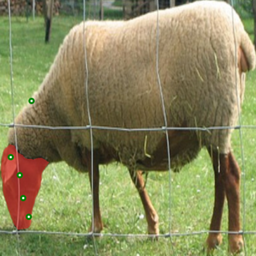}
    \end{subfigure}
    \begin{subfigure}[b]{0.24\textwidth}
        \centering
        \includegraphics[width=\textwidth, height = 2.8cm]{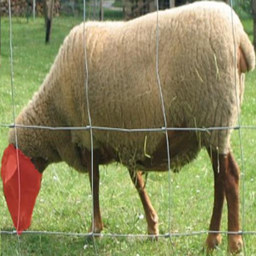}
    \end{subfigure}
    \\
    
    \begin{subfigure}[b]{0.24\textwidth}
        \centering
        \includegraphics[width=\textwidth, height = 2.8cm]{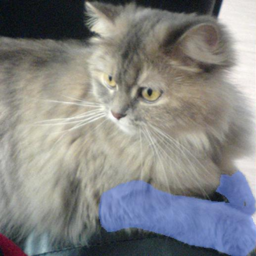}
        \caption{Support Image}     
    \end{subfigure}
    \begin{subfigure}[b]{0.24\textwidth}
        \centering
        \includegraphics[width=\textwidth, height = 2.8cm]{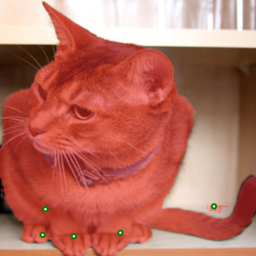}
        \caption{Baseline}     
    \end{subfigure}
    \begin{subfigure}[b]{0.24\textwidth}
        \centering
        \includegraphics[width=\textwidth, height = 2.8cm]{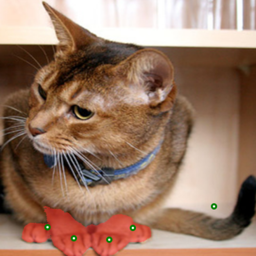}
        \caption{Refined (Ours)}     
    \end{subfigure}
    \begin{subfigure}[b]{0.24\textwidth}
        \centering
        \includegraphics[width=\textwidth, height = 2.8cm]{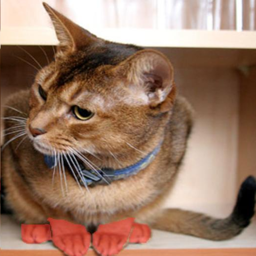}
        \caption{Ground Truth}     
    \end{subfigure}
    \caption{Qualitative result of \ourmodel{} with Matcher in part segmentation.}
        \vspace{-15pt}
\end{figure}

\newpage
\subsection{Five shot semantic segmentation (Matcher)}
\label{sec:5semantic}

\begin{figure}[h!]
    
    \centering
    \begin{subfigure}[b]{0.18\textwidth}
        \centering
        \includegraphics[width=\textwidth, height = 3.2cm]{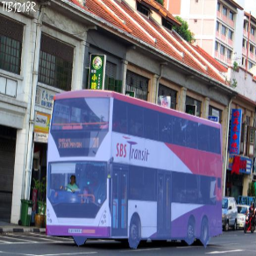}
    \end{subfigure}
    \begin{subfigure}[b]{0.18\textwidth}
        \centering
        \includegraphics[width=\textwidth, height = 3.2cm]{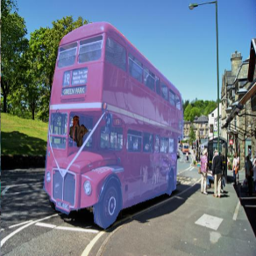}
    \end{subfigure}
    \begin{subfigure}[b]{0.18\textwidth}
        \centering
        \includegraphics[width=\textwidth, height = 3.2cm]{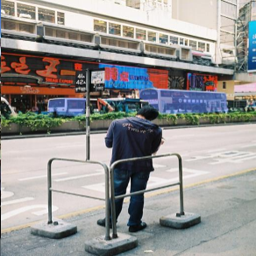}
    \end{subfigure}
    \begin{subfigure}[b]{0.18\textwidth}
        \centering
        \includegraphics[width=\textwidth, height = 3.2cm]{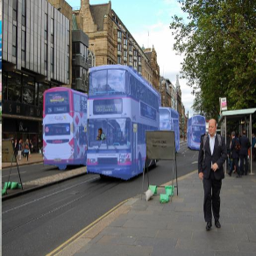}
    \end{subfigure}
    \begin{subfigure}[b]{0.18\textwidth}
        \centering
        \includegraphics[width=\textwidth, height = 3.2cm]{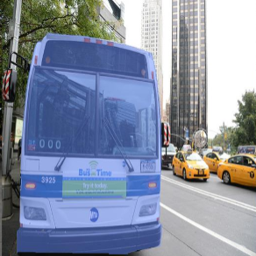}
    \end{subfigure}
    \\
    \begin{subfigure}[b]{0.18\textwidth}
        \centering
        \includegraphics[width=\textwidth, height = 3.2cm]{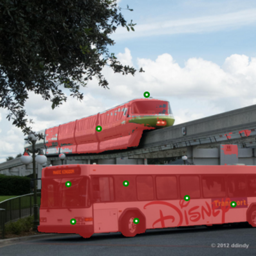}
    \end{subfigure}
    \begin{subfigure}[b]{0.18\textwidth}
        \centering
        \includegraphics[width=\textwidth, height = 3.2cm]{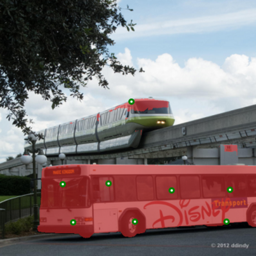}
    \end{subfigure}
    \begin{subfigure}[b]{0.18\textwidth}
        \centering
        \includegraphics[width=\textwidth, height = 3.2cm]{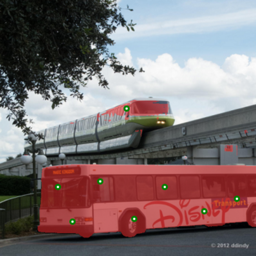}
    \end{subfigure}
    \begin{subfigure}[b]{0.18\textwidth}
        \centering
        \includegraphics[width=\textwidth, height = 3.2cm]{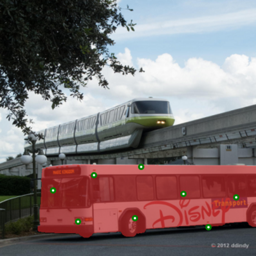}
    \end{subfigure}
    \begin{subfigure}[b]{0.18\textwidth}
        \centering
        \includegraphics[width=\textwidth, height = 3.2cm]{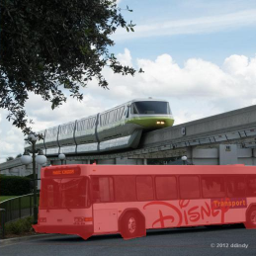}
    \end{subfigure}
    
    \vspace{-0.5em}
    \rule{0.95\textwidth}{0.4pt}
    \vspace{0.5em}
    
    \begin{subfigure}[b]{0.18\textwidth}
        \centering
        \includegraphics[width=\textwidth, height = 3.2cm]{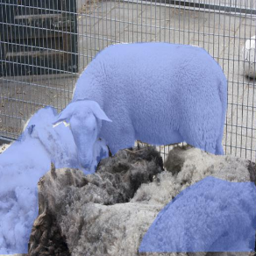}
    \end{subfigure}
    \begin{subfigure}[b]{0.18\textwidth}
        \centering
        \includegraphics[width=\textwidth, height = 3.2cm]{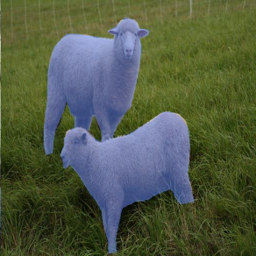}
    \end{subfigure}
    \begin{subfigure}[b]{0.18\textwidth}
        \centering
        \includegraphics[width=\textwidth, height = 3.2cm]{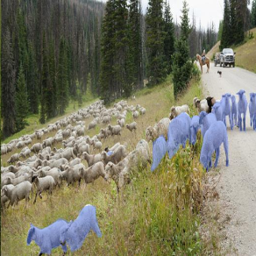}
    \end{subfigure}
    \begin{subfigure}[b]{0.18\textwidth}
        \centering
        \includegraphics[width=\textwidth, height = 3.2cm]{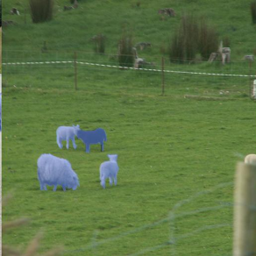}
    \end{subfigure}
    \begin{subfigure}[b]{0.18\textwidth}
        \centering
        \includegraphics[width=\textwidth, height = 3.2cm]{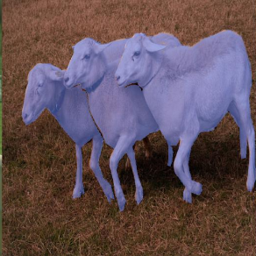}
    \end{subfigure}
    \\
    \begin{subfigure}[b]{0.18\textwidth}
        \centering
        \includegraphics[width=\textwidth, height = 3.2cm]{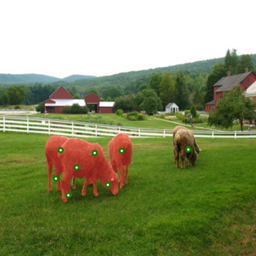}
    \end{subfigure}
    \begin{subfigure}[b]{0.18\textwidth}
        \centering
        \includegraphics[width=\textwidth, height = 3.2cm]{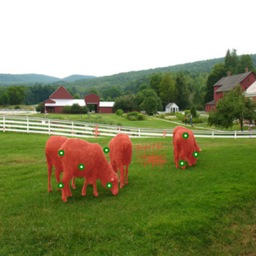}
    \end{subfigure}
    \begin{subfigure}[b]{0.18\textwidth}
        \centering
        \includegraphics[width=\textwidth, height = 3.2cm]{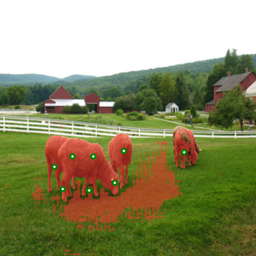}
    \end{subfigure}
    \begin{subfigure}[b]{0.18\textwidth}
        \centering
        \includegraphics[width=\textwidth, height = 3.2cm]{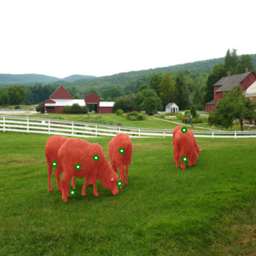}
    \end{subfigure}
    \begin{subfigure}[b]{0.18\textwidth}
        \centering
        \includegraphics[width=\textwidth, height = 3.2cm]{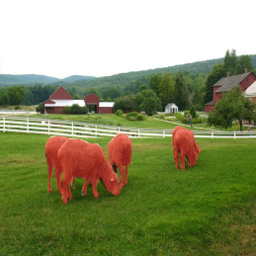}
    \end{subfigure}
    
    \vspace{-0.5em}
    \rule{0.95\textwidth}{0.4pt}
    \vspace{0.5em}

    \begin{subfigure}[b]{0.18\textwidth}
        \centering
        \includegraphics[width=\textwidth, height = 3.2cm]{figures/appendix/semantic/5shot-matcher/coco2-2/ref1.png}
    \end{subfigure}
    \begin{subfigure}[b]{0.18\textwidth}
        \centering
        \includegraphics[width=\textwidth, height = 3.2cm]{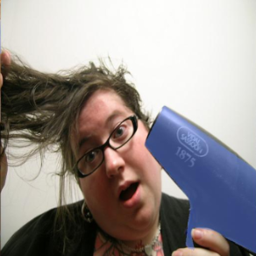}
    \end{subfigure}
    \begin{subfigure}[b]{0.18\textwidth}
        \centering
        \includegraphics[width=\textwidth, height = 3.2cm]{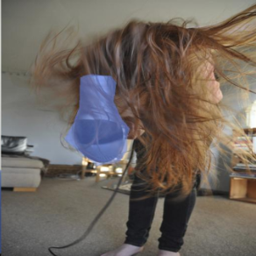}
    \end{subfigure}
    \begin{subfigure}[b]{0.18\textwidth}
        \centering
        \includegraphics[width=\textwidth, height = 3.2cm]{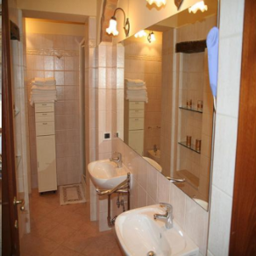}
    \end{subfigure}
    \begin{subfigure}[b]{0.18\textwidth}
        \centering
        \includegraphics[width=\textwidth, height = 3.2cm]{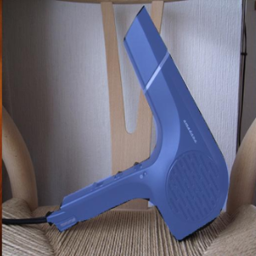}
    \end{subfigure}
    \\
    \begin{subfigure}[b]{0.18\textwidth}
        \centering
        \includegraphics[width=\textwidth, height = 3.2cm]{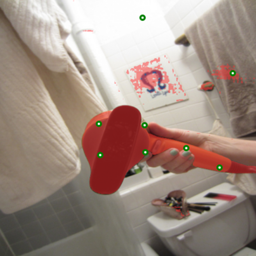}
    \end{subfigure}
    \begin{subfigure}[b]{0.18\textwidth}
        \centering
        \includegraphics[width=\textwidth, height = 3.2cm]{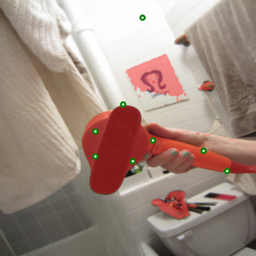}
    \end{subfigure}
    \begin{subfigure}[b]{0.18\textwidth}
        \centering
        \includegraphics[width=\textwidth, height = 3.2cm]{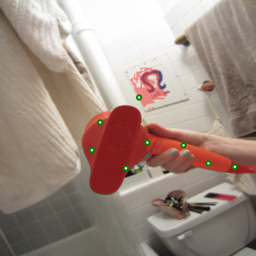}
    \end{subfigure}
    \begin{subfigure}[b]{0.18\textwidth}
        \centering
        \includegraphics[width=\textwidth, height = 3.2cm]{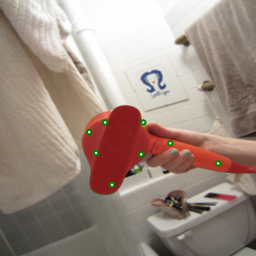}
    \end{subfigure}
    \begin{subfigure}[b]{0.18\textwidth}
        \centering
        \includegraphics[width=\textwidth, height = 3.2cm]{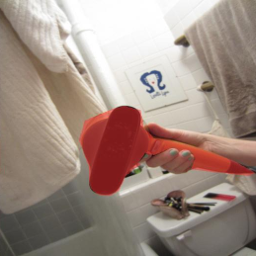}
    \end{subfigure}
\caption{Qualitative result of five shot semantic segmentation (Matcher). The test samples are separated by dividers. For each sample, the upper row shows the support images, while the lower row displays, from left to right, the baseline result, the result of \ourmodel{} across iterations, and the ground truth.}
\vspace{-12pt}
\end{figure}
\vfill
\newpage
\subsection{Five shot part segmentation (Matcher)}

\label{sec:5part}

\begin{figure}[h!]
    \centering
    \begin{subfigure}[b]{0.18\textwidth}
        \centering
        \includegraphics[width=\textwidth, height = 3.2cm]{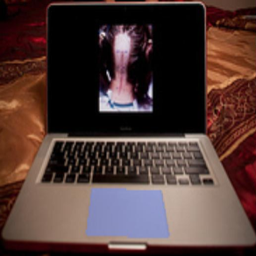}
    \end{subfigure}
    \begin{subfigure}[b]{0.18\textwidth}
        \centering
        \includegraphics[width=\textwidth, height = 3.2cm]{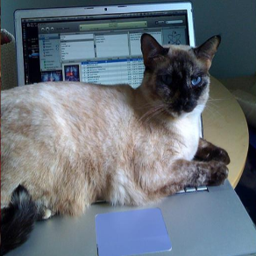}
    \end{subfigure}
    \begin{subfigure}[b]{0.18\textwidth}
        \centering
        \includegraphics[width=\textwidth, height = 3.2cm]{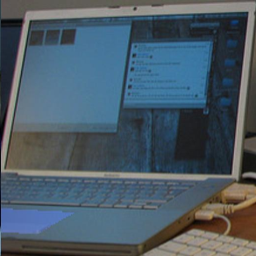}
    \end{subfigure}
    \begin{subfigure}[b]{0.18\textwidth}
        \centering
        \includegraphics[width=\textwidth, height = 3.2cm]{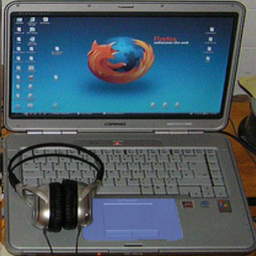}
    \end{subfigure}
    \begin{subfigure}[b]{0.18\textwidth}
        \centering
        \includegraphics[width=\textwidth, height = 3.2cm]{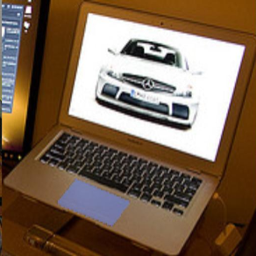}
    \end{subfigure}
    \\
    \begin{subfigure}[b]{0.18\textwidth}
        \centering
        \includegraphics[width=\textwidth, height = 3.2cm]{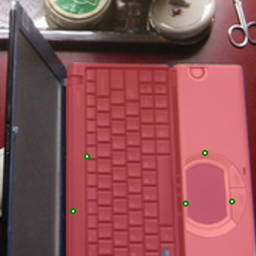}
    \end{subfigure}
    \begin{subfigure}[b]{0.18\textwidth}
        \centering
        \includegraphics[width=\textwidth, height = 3.2cm]{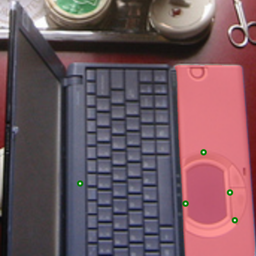}
    \end{subfigure}
    \begin{subfigure}[b]{0.18\textwidth}
        \centering
        \includegraphics[width=\textwidth, height = 3.2cm]{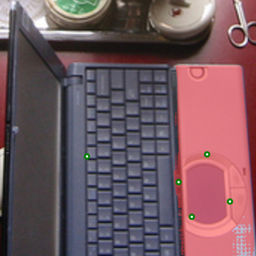}
    \end{subfigure}
    \begin{subfigure}[b]{0.18\textwidth}
        \centering
        \includegraphics[width=\textwidth, height = 3.2cm]{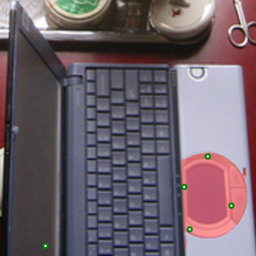}
    \end{subfigure}
    \begin{subfigure}[b]{0.18\textwidth}
        \centering
        \includegraphics[width=\textwidth, height = 3.2cm]{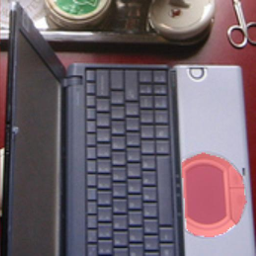}
    \end{subfigure}
    
    \vspace{-0.5em}
    \rule{0.95\textwidth}{0.4pt}
    \vspace{0.5em}
    
    \begin{subfigure}[b]{0.18\textwidth}
        \centering
        \includegraphics[width=\textwidth, height = 3.2cm]{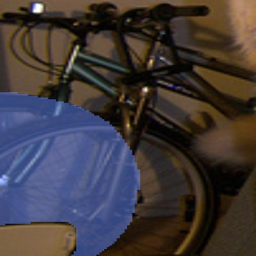}
    \end{subfigure}
    \begin{subfigure}[b]{0.18\textwidth}
        \centering
        \includegraphics[width=\textwidth, height = 3.2cm]{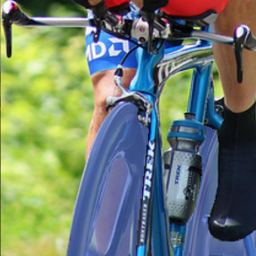}
    \end{subfigure}
    \begin{subfigure}[b]{0.18\textwidth}
        \centering
        \includegraphics[width=\textwidth, height = 3.2cm]{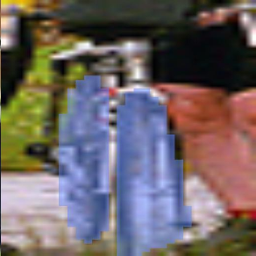}
    \end{subfigure}
    \begin{subfigure}[b]{0.18\textwidth}
        \centering
        \includegraphics[width=\textwidth, height = 3.2cm]{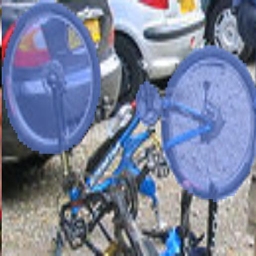}
    \end{subfigure}
    \begin{subfigure}[b]{0.18\textwidth}
        \centering
        \includegraphics[width=\textwidth, height = 3.2cm]{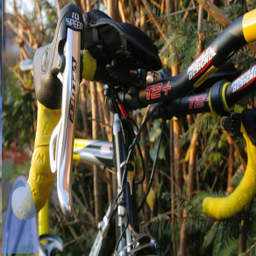}
    \end{subfigure}
    \\
    \begin{subfigure}[b]{0.18\textwidth}
        \centering
        \includegraphics[width=\textwidth, height = 3.2cm]{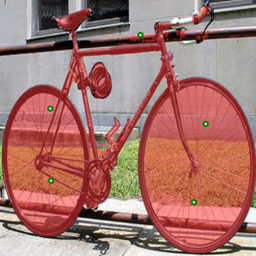}
    \end{subfigure}
    \begin{subfigure}[b]{0.18\textwidth}
        \centering
        \includegraphics[width=\textwidth, height = 3.2cm]{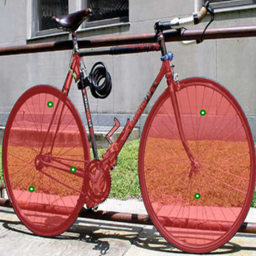}
    \end{subfigure}
    \begin{subfigure}[b]{0.18\textwidth}
        \centering
        \includegraphics[width=\textwidth, height = 3.2cm]{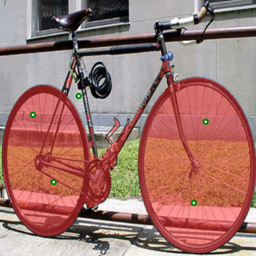}
    \end{subfigure}
    \begin{subfigure}[b]{0.18\textwidth}
        \centering
        \includegraphics[width=\textwidth, height = 3.2cm]{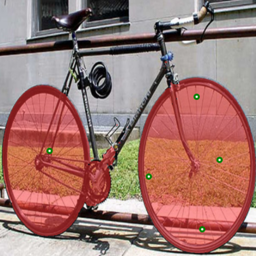}
    \end{subfigure}
    \begin{subfigure}[b]{0.18\textwidth}
        \centering
        \includegraphics[width=\textwidth, height = 3.2cm]{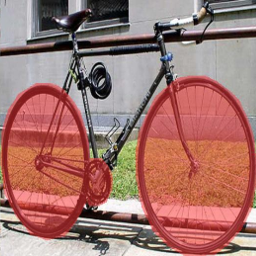}
    \end{subfigure}
    
    \vspace{-0.5em}
    \rule{0.95\textwidth}{0.4pt}
    \vspace{0.5em}

    \begin{subfigure}[b]{0.18\textwidth}
        \centering
        \includegraphics[width=\textwidth, height = 3.2cm]{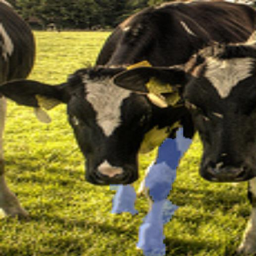}
    \end{subfigure}
    \begin{subfigure}[b]{0.18\textwidth}
        \centering
        \includegraphics[width=\textwidth, height = 3.2cm]{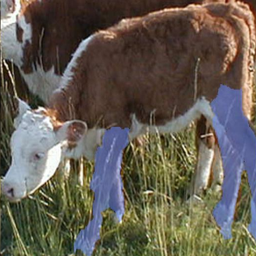}
    \end{subfigure}
    \begin{subfigure}[b]{0.18\textwidth}
        \centering
        \includegraphics[width=\textwidth, height = 3.2cm]{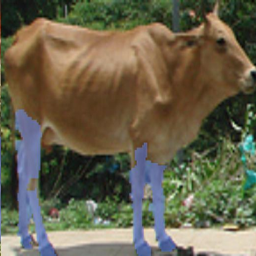}
    \end{subfigure}
    \begin{subfigure}[b]{0.18\textwidth}
        \centering
        \includegraphics[width=\textwidth, height = 3.2cm]{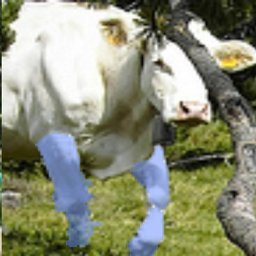}
    \end{subfigure}
    \begin{subfigure}[b]{0.18\textwidth}
        \centering
        \includegraphics[width=\textwidth, height = 3.2cm]{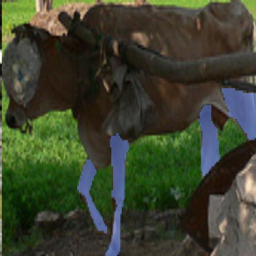}
    \end{subfigure}
    \\
    \begin{subfigure}[b]{0.18\textwidth}
        \centering
        \includegraphics[width=\textwidth, height = 3.2cm]{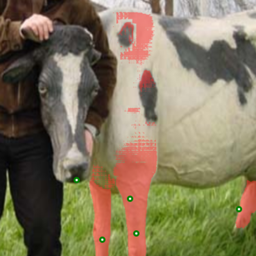}
    \end{subfigure}
    \begin{subfigure}[b]{0.18\textwidth}
        \centering
        \includegraphics[width=\textwidth, height = 3.2cm]{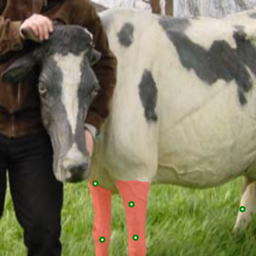}
    \end{subfigure}
    \begin{subfigure}[b]{0.18\textwidth}
        \centering
        \includegraphics[width=\textwidth, height = 3.2cm]{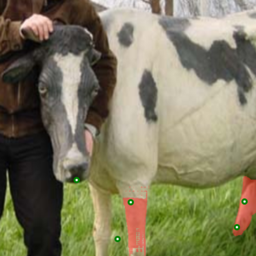}
    \end{subfigure}
    \begin{subfigure}[b]{0.18\textwidth}
        \centering
        \includegraphics[width=\textwidth, height = 3.2cm]{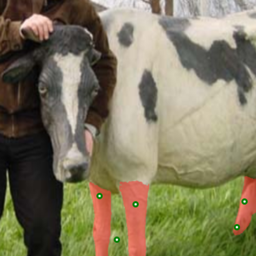}
    \end{subfigure}
    \begin{subfigure}[b]{0.18\textwidth}
        \centering
        \includegraphics[width=\textwidth, height = 3.2cm]{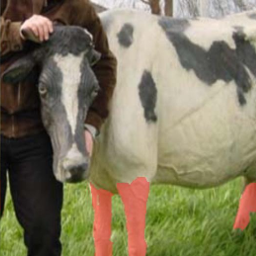}
    \end{subfigure}
\caption{Qualitative result of five shot part segmentation (Matcher). The test samples are separated by dividers. For each sample, the upper row shows the support images, while the lower row displays, from left to right, the baseline result, the result of \ourmodel{} across iterations, and the ground truth.}        
\vspace{-12pt}
\end{figure}
\vfill

\newpage
\subsection{Comparison of the effects of \ourmodel{} across different baselines.}
\begin{figure}[h!]
    \centering
    \begin{subfigure}[b]{0.2\textwidth}
        \centering
        \includegraphics[width=\textwidth, height=3.2cm]{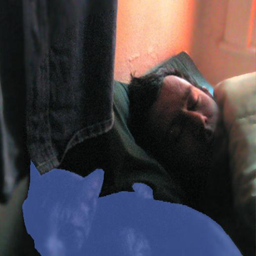}
        \caption{Support Image}
    \end{subfigure}
    \begin{subfigure}[b]{0.2\textwidth}
        \centering
        \includegraphics[width=\textwidth, height=3.2cm]{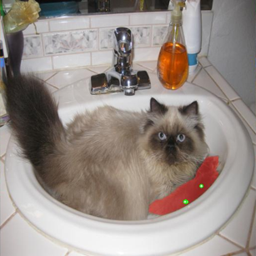}
        \caption{PerSAM-F}
        \label{fig:9.b}
    \end{subfigure}
    \begin{subfigure}[b]{0.2\textwidth}
        \centering
        \includegraphics[width=\textwidth, height=3.2cm]{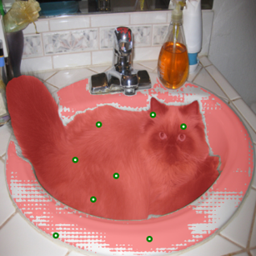}
        \caption{Matcher}
        \label{fig:9.d}
    \end{subfigure}

    \begin{subfigure}[b]{0.2\textwidth}
        \centering
        \includegraphics[width=\textwidth, height=3.2cm]{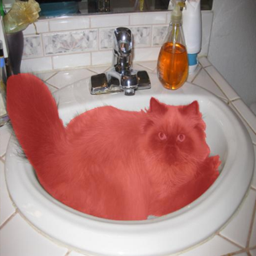}
        \caption{Ground Truth}
    \end{subfigure}
    \begin{subfigure}[b]{0.2\textwidth}
        \centering
        \includegraphics[width=\textwidth, height=3.2cm]{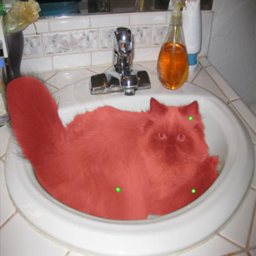}
        \caption{PerSAM-F+\ourmodel{}}
        \label{fig:9.g}
    \end{subfigure}
    \begin{subfigure}[b]{0.2\textwidth}
        \centering
        \includegraphics[width=\textwidth, height=3.2cm]{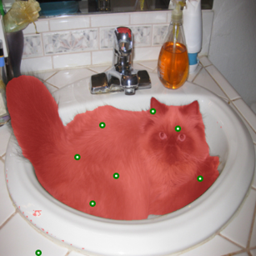}
        \caption{Matcher+\ourmodel{}}
        \label{fig:9.i}
    \end{subfigure}

    \caption{Illustration of prompt refinement of \ourmodel{}}
    \label{fig:9}
\end{figure}

\large{
Comparing the impact of \ourmodel{} across different baselines on the same image, as shown in \cref{fig:9}, provides a more comprehensive understanding of our method's effectiveness. For PerSAM-F (\cref{fig:9.b,fig:9.g}), prompts are selected by top-k sampling. \ourmodel{} addresses the issue of congested prompts, resolving object ambiguity. For Matcher (\cref{fig:9.d,fig:9.i}), the robust sampler distributes the prompts more widely. However, \ourmodel{} further fine-tunes them, enhancing precision and producing more accurate segmentation masks.}

\newpage
\section{Sensitivity analysis of $\gamma$ and $\eta$ across object categories.}


\begin{table}[!h]
\centering
\small
\vspace{-5pt}
\resizebox{0.8\textwidth}{!}{%
\begin{tabular}{c|c|c|ccc|ccc|ccc}
\hline
 &  &  & \multicolumn{3}{c|}{$\eta_1=1e^{-2}$} 
     & \multicolumn{3}{c|}{$\eta_2=1e^{-3}$}
     & \multicolumn{3}{c}{$\eta_3=1e^{-4}$} \\ 
$c_i$ & Metric & Base$^{*}$ 
& $\gamma_1$ & $\gamma_2$ & $\gamma_3$
& $\gamma_1$ & $\gamma_2$ & $\gamma_3$
& $\gamma_1$ & $\gamma_2$ & $\gamma_3$ \\
\hline

\multirow{2}{*}{$c_1$}
& Mean & 55.02
& 58.64 & 70.32 & 64.18
& 65.20 & 62.88 & 63.77
& 62.90 & 63.77 & 58.48 \\
& Std  & (30.65)
& (24.37) & (27.82) & (28.14)
& (26.94) & (27.90) & (27.89)
& (27.87) & (27.89) & (27.51) \\
\hline

\multirow{2}{*}{$c_2$}
& Mean & 40.34
& 52.95 & 56.67 & 50.19
& 50.13 & 46.96 & 45.62
& 46.96 & 48.54 & 48.64 \\
& Std  & (38.56)
& (33.21) & (37.06) & (37.36)
& (34.19) & (36.87) & (37.56)
& (36.87) & (38.20) & (38.26) \\
\hline

\multirow{2}{*}{$c_3$}
& Mean & 45.89
& 79.55 & 70.07 & 57.45
& 69.37 & 56.56 & 54.41
& 54.47 & 54.41 & 53.90 \\
& Std  & (33.48)
& (24.49) & (28.62) & (29.46)
& (28.16) & (31.82) & (31.94)
& (32.49) & (31.94) & (32.83) \\
\hline

\multirow{2}{*}{$c_4$}
& Mean & 64.14
& 74.22 & 71.94 & 67.82
& 72.86 & 67.63 & 68.55
& 67.49 & 68.55 & 70.68 \\
& Std  & (29.81)
& (23.77) & (27.50) & (24.99)
& (24.41) & (24.80) & (26.26)
& (24.73) & (26.26) & (25.00) \\
\hline

\multirow{2}{*}{$c_5$}
& Mean & 73.16
& 73.74 & 80.14 & 74.46
& 74.24 & 74.23 & 84.07
& 74.25 & 74.88 & 78.56 \\
& Std  & (27.60)
& (25.10) & (24.29) & (22.70)
& (26.55) & (26.46) & (17.85)
& (26.46) & (20.93) & (22.42) \\
\hline

\multirow{2}{*}{$c_6$}
& Mean & 92.51
& 89.62 & 95.99 & 89.70
& 95.99 & 89.70 & 90.39
& 89.71 & 90.39 & 90.76 \\
& Std  & (10.59)
& (19.79) & (0.97) & (13.40)
& (0.97) & (13.40) & (17.28)
& (13.41) & (17.28) & (16.02) \\
\hline

\multirow{2}{*}{$c_7$}
& Mean & 35.94
& 72.44 & 72.82 & 59.56
& 74.76 & 63.74 & 46.88
& 63.74 & 49.13 & 47.00 \\
& Std  & (31.90)
& (28.90) & (23.68) & (29.49)
& (20.44) & (20.72) & (26.31)
& (20.73) & (25.40) & (28.95) \\
\hline

\multirow{2}{*}{$c_8$}
& Mean & 49.35
& 72.80 & 71.34 & 66.13
& 71.34 & 70.63 & 65.43
& 70.63 & 64.04 & 53.30 \\
& Std  & (27.27)
& (21.90) & (22.20) & (25.94)
& (22.20) & (22.83) & (28.07)
& (22.83) & (28.15) & (27.41) \\
\hline

\multirow{2}{*}{$c_9$}
& Mean & 72.78
& 77.23 & 69.07 & 74.76
& 67.77 & 75.48 & 64.41
& 74.85 & 64.42 & 64.48 \\
& Std  & (33.33)
& (31.84) & (34.11) & (31.18)
& (35.45) & (31.19) & (36.73)
& (31.06) & (36.73) & (36.62) \\
\hline
\hline
\addlinespace[2pt]
\multicolumn{12}{l}{\large
$\gamma$-Avg (avg. over $\eta$ and classes):\;
$\gamma_1{=}70.29,\; \gamma_2{=}68.31,\; \gamma_3{=}64.95$} \\
\multicolumn{12}{l}{\large
$\eta$-Avg (avg. over $\gamma$ and classes):\;
$\eta_1{=}70.88,\; \eta_2{=}67.89,\; \eta_3{=}64.78$} \\
\addlinespace[2pt]
\hline

\end{tabular}
}
\vspace{-5pt}
\caption{\footnotesize Overall mIoU(\%) results of PR-MaGIC across $\eta$, $\gamma$ and classes ($c_i$) configurations.
Base$^{*}$: PerSAM-F. Each $c_i$ includes 10 samples per class:
$\{\texttt{gas\_pump}, \texttt{indian\_cobra}, \texttt{cocktail\_shaker},
\texttt{golfcart}, \texttt{teddy}, \texttt{garbage\_truck},
\texttt{stapler}, \texttt{children\_slide}, \texttt{studio\_couch}\}$.
$\gamma_1 = 1$, $\gamma_2 = 10^{-1}$, $\gamma_3 = 10^{-2}$.}
\label{tab:mean_std_two_rows}
\end{table}

As shown in \cref{tab:mean_std_two_rows}, sensitivity analysis on $\gamma$ and $\eta$ across randomly selected 9 classes was conducted. PR-MaGIC consistently improves mIoU across most classes regardless of hyperparameter settings, demonstrating class-agnostic gains despite minor exceptions in $c_6$ and $c_9$. The $\eta$-averaged results ($\eta_1{=}70.88 > \eta_2{=}67.89 > \eta_3{=}64.78$) show that larger step sizes generally lead to better performance, suggesting that the refinement gradients are well aligned with the intended update direction. Likewise, the $\gamma$-averaged trend ($\gamma_1{=}70.29 > \gamma_2{=}68.31 > \gamma_3{=}64.95$) indicates that stronger entropy regularization results in more robust refinement behavior.

As our method targets test-time scaling, we adopt a conservative and minimal hyperparameter tuning strategy to assess overall performance gains. While the additional analysis in this appendix is conducted on a limited subset, the configuration used in our main experiments already achieves strong and consistent results, with other settings yielding even higher performance in some cases. We expect that this class-wise tuning would be valuable for practical deployment.

\end{document}